\documentclass[lettersize,journal]{IEEEtran}
\usepackage{amsmath,amsfonts}
\usepackage{algorithmic}
\usepackage{algorithm}
\usepackage{array}
\usepackage[caption=false,font=normalsize,labelfont=sf,textfont=sf]{subfig}
\usepackage{textcomp}
\usepackage{stfloats}
\usepackage{url}
\usepackage{verbatim}
\usepackage{graphicx}
\usepackage{cite}
\usepackage[breaklinks=true,colorlinks,bookmarks=false]{hyperref}

\begin{document}

\title{Ins-HOI: Instance Aware Human-Object Interactions Recovery}

\author{Jiajun Zhang,
        Yuxiang Zhang,
        Hongwen Zhang,
        Xiao Zhou,
        Boyao Zhou,
        Ruizhi Shao,
        Zonghai Hu, 
        \\
        and Yebin Liu,~\IEEEmembership{Member,~IEEE}
\IEEEcompsocitemizethanks{
\IEEEcompsocthanksitem Jiajun Zhang, Zonghai Hu are with the School of Electronic Engineering, Beijing University of Posts and Telecommunications, Beijing 100876, China. E-mail: \{jiajun.zhang, zhhu\}@bupt.edu.cn
\IEEEcompsocthanksitem Yuxiang Zhang, Xiao Zhou, Boyao Zhou, Ruizhi Shao, and Yebin Liu are with the Department of Automation, Tsinghua University, Beijing 100084, China. E-mail: \{yx-z19, shaorz20\}@mails.tsinghua.edu.cn; xzhou@tsinghua.edu.cn; \{bzhou22, liuyebin\}@mail.tsinghua.edu.cn 
\IEEEcompsocthanksitem Hongwen Zhang is with the School of Artificial Intelligence, Beijing Normal University, Beijing 100875, China. E-mail: zhanghongwen@bnu.edu.cn
\IEEEcompsocthanksitem Corresponding author: Yebin Liu
}}

\markboth{Journal of \LaTeX\ Class Files,~Vol.~14, No.~8, August~2021}%
{Shell \MakeLowercase{\textit{et al.}}: A Sample Article Using IEEEtran.cls for IEEE Journals}

\maketitle

\begin{abstract}
Accurately modeling detailed interactions between human/hand and object is an appealing yet challenging task. 
Current multi-view capture systems are only capable of reconstructing multiple subjects into a single, unified mesh, which fails to model the states of each instance individually during interactions. To address this, previous methods use template-based representations to track human/hand and object. 
However, the quality of the reconstructions is limited by the descriptive capabilities of the templates so that these methods are inherently struggle with geometry details, pressing deformations and invisible contact surfaces.
In this work, we propose an end-to-end \textbf{Ins}tance-aware \textbf{H}uman-\textbf{O}bject \textbf{I}nteractions recovery (Ins-HOI) framework by introducing an instance-level occupancy field representation.
However, the real-captured data is presented as a holistic mesh, unable to provide instance-level supervision.
To address this, we further propose a complementary training strategy that leverages synthetic data to introduce instance-level shape priors, enabling the disentanglement of occupancy fields for different instances.
Specifically, synthetic data, created by randomly combining individual scans of humans/hands and objects, guides the network to learn a coarse prior of instances. Meanwhile, real-captured data helps in learning the overall geometry and restricting interpenetration in contact areas. 
As demonstrated in experiments, our method Ins-HOI supports instance-level reconstruction and provides reasonable and realistic invisible contact surfaces even in cases of extremely close interaction. 
To facilitate the research of this task, we collect a large-scale, high-fidelity 3D scan dataset, including 5.2k high-quality scans with real-world human-chair and hand-object interactions. 
The code and data will be public for research purposes. 
Data examples and the video results of our method can be found at: 
\href{https://jiajunzhang16.github.io/ins-hoi}{https://jiajunzhang16.github.io/ins-hoi}.

\end{abstract}

\section{Introduction}\label{sec:introduction}

\IEEEPARstart{H}{uman/hand} object interactions (HOI) have significant applications in the fields of computer vision, robotics, and virtual reality.
Accurately reconstructing the state of interaction enables a comprehensive understanding, analysis, and generation of human behavior. 
There is remarkable progress in separately reconstructing human~\cite{pifuSHNMKL19, zheng2020pamir, He2021ARCHAC,xiu2022icon, xiu2023econ}, hand~\cite{Zhang2021twohand, Park_2022_CVPR_HandOccNet, Li2022intaghand, yu2023acr}, and objects~\cite{Park_2019_deepsdf, zheng2021deep, Boulch_2022_CVPR_poco, jiang2020sdfdiff, wang2023alto}.
However, these methods are designed to reconstruct individual object and do not support the simultaneous reconstruction of two interacting instances.
To overcome this limitation, numerous datasets and reconstruction methods have been specifically developed for human-object~\cite{PROX:2019, bhatnagar22behave, zhang2022couch, xu2021d3dhoi, jiang2023chairs, zhang2023neuraldome}, and hand-object~\cite{ GRAB:2020,  fan2023arctic, YangCVPR2022OakInk, Li_2023_ICCV_CHORD, hasson19_obman, GraspingField:3DV:2020, hampali2020honnotate, huang2022hhor, ye2022hand} interactions. While these methods achieve reasonable results, they struggle to capture the intricate geometry and plausible invisible contact surfaces.

\begin{figure}
  \centering
  \includegraphics[width=1.0\linewidth]{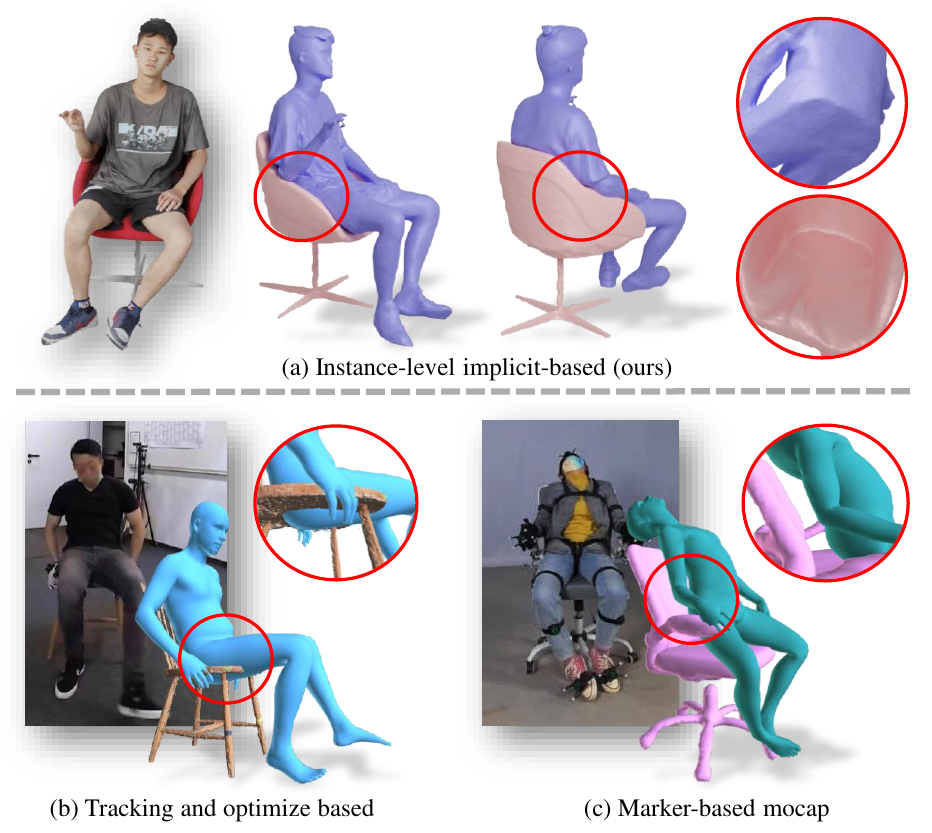}
  \caption{Our instance-level implicit-based approach achieves accurate reconstruction of the geometry and invisible contact areas. In contrast, tracking and optimization-based methods~\cite{bhatnagar22behave}, as well as marker-based methods~\cite{jiang2023chairs}, rely on human parametric template and present inaccuracies in interaction region and lack fine-detailed geometry.}
  \label{fig:teaser}
\end{figure}

A major limitation of previous HOI recovery solutions is that the template-based representations are employed to track the human/hand and objects in interactions.
For instance, the parametric models like SMPL~\cite{SMPL:2015,SMPL-X:2019} and MANO~\cite{MANO:SIGGRAPHASIA:2017} are used to represent the human/hand poses, while pre-defined templates are selected as the object representations. Constrained by the expressive capacity of template-based representations, these methods exhibit limited capabilities as they are unable to reconstruct clothing or the fine deformations caused by contact, nor can they address issues of model penetration and gap, as shown in Fig.~\ref{fig:teaser}. Regarding the interacting objects, most previous works assume the objects to be rigid, reconstructing them by simply tracking the pose of object templates~\cite{zhang2023neuraldome, zhang2020phosa, hampali2020honnotate}. This simplification of human/hand and object representations falls short in effectively handling the contact surfaces, which are crucial for accurately reconstructing close interactions.

Diverging from previous template-based solutions, we would argue for a more suitable HOI modeling paradigm by introducing instance-level human/hand-object reconstruction via implicit surface representations. The implicit representations \cite{implicit} are characterized by their continuous property, high-resolution details, and robustness to topological variations, which is more suitable to model the intricate geometric appearance and invisible contact surfaces of the two interacting instances. 
Based on implicit representations, we present Ins-HOI, an end-to-end approach for human/hand-object interaction recovery at the instance level.
As previous methods~\cite{OccNet:CVPR19,pifuSHNMKL19} typically reconstruct the target scene as a single entity, the existing implicit field can not model distinct instances simultaneously.
To handle this issue, we extend the implicit function by proposing an instance-level occupancy field to learn separate geometry for each instance.
Such an extension not only allows our representation to support instance-aware perception but also enables a more direct constraint between the contact surfaces of different instances in our algorithm.
Such a representation also allow our method to learn contact priors implicitly from sparse multi-view observations.

Though our representation is well-suited for instance-level reconstruction, achieving this goal is still non-trivial as it is challenging to acquire individual instance-level ground truth for two tightly interacting objects in real-world captured data.
To address this issue, we propose a complementary training strategy to enhance the learning of individual geometric priors and the perception of interaction. 
Alongside real-captured data, we augment our training data with synthesized data. These are created by randomly composing human/hand scans with object scans, and intentionally allowing for some extent of penetration.
These synthesized data have their ground truth instance shape (penetration areas are assigned to both human/hand and object) to make the network better learning the individual shape as completely as possible even under extremely closely interacting and occluded scenarios. For the real scans, it serves as a penalization to make our instance reconstructions share no penetration and assure seamless integration with the original real scans.
Overall, the two types of data play adversarial and complementary roles in the training process. The synthesized data focuses on the completeness of individual shapes, while the real-captured scans concentrate more on overall reasonableness.

In this paper, we demonstrate that Ins-HOI can handle two of the most common interaction scenarios: human-chair and hand-object interactions.
Compared with previous methods~\cite{pifuSHNMKL19, neus2}, Ins-HOI achieves comparable reconstruction results when considering the interacting human/hand and objects as a whole. Besides, Ins-HOI supports instance-level reconstruction, enabling the precise separation and detailed modeling of individual object instances within a scene.
In addition, Ins-HOI can produce reasonable and realistic invisible contact surfaces even in cases of extremely close interaction, which are not annotated in the real-scanned dataset.
Our method requires only one forward pass, eliminating the need for tracking, modeling, and registration processes typically required by other HOI reconstruction methods~\cite{bhatnagar22behave, yin2023hi4d, zhang2023neuraldome, jiang2023chairs, hampali2020honnotate}. 
We hope our proposed method and datasets for instance-level human/hand-object recovery can pave the way for new research direction.

The contributions of this work can be summarized as follows:
\begin{itemize}
    \item We propose instance-level human/hand-object reconstruction via implicit surface representations for the task of human/hand-object interaction. Our method learns the contact prior implicitly and produces reasonable and realistic invisible contact surfaces even in cases of extremely close interaction.
    \item We propose a complementary training strategy to tackle the lack of instance-level ground truths in real-scanned datasets. Such a strategy can well balance the completeness of individual shapes and the overall reasonableness of invisible contact regions.
    \item A large-scale, high-fidelity 3D scan dataset of human/hand-object interactions, named Ins-Sit and Ins-Grasp, comprising a total of 5.2k scans. We benchmark the task of instance-level reconstruction on our dataset. 
\end{itemize}

\section{Related Work}\label{sec:related_work}

\subsection{Human Object Interactions}
Recent studies on modeling human-object interactions can generally be summarized into two categories. The first category is tracking-based approaches~\cite{bhatnagar22behave, huang2022intercap, jiang2023chairs, zhang2022couch, PROX:2019, GRAB:2020, li2023object, zhang2020phosa}, where motion sequences of human-object interactions are captured through a multi-view setup, with RGB or RGBD inputs. They utilize motion capture systems or optimization methods to obtain parametric representations of the human body~\cite{SMPL:2015, SMPL-X:2019} and the pose of objects. During the post-processing phase, they incorporate various constraints on contact and penetration to refine the results. Thanks to their efforts in data collection, numerous studies on human-object interaction, such as analyzing contact~\cite{shimada2022hulc, chen2023hot, zhang2021learninglemo, hassan2021populating}, spatial arrangement~\cite{zhang2020phosa, xie2022chore, xie2023vistracker, wang2022reconstructing} and generation~\cite{summon, petrov2023popup, kulkarni2023nifty, xu2023interdiff, Pi_2023_ICCV} have significantly advanced this field. These methods represent the human form using parametric models and based on the assumption that object is given. We argue that the ideal representation of the human would have detailed geometry rather than relying on parametric model, which fall short in simulating the clothes, the deformation of muscle tissue under pressure. Likewise, objects represented by pre-obtained template cannot effectively model the deformations caused by contact and lack of generalization. The second category seeks to directly reconstruct the geometry of two interacting instances~\cite{zhang2023neuraldome, yin2023hi4d} instead of representing them with parametric models, which is similar to our works. NeuralDome~\cite{zhang2023neuraldome} introduce a neural pipeline to accurate modeling human and object. Despite achieving impressive results, two issues still remain. The first issue is the heavy setup, their capture system relies on 76 camera viewpoints and 16 optical motion capture devices, which is hard to deploy for lab-external environments. Additionally, as a multi-stage method, it requires independent modeling~\cite{weng_humannerf_2022_cvpr, zhang2021editable} for each subject, increasing the complexity of the implementation process. The second issue is that they still based on given object templates and track the object pose, which means the non-rigid deformations caused by contact cannot be effectively modeled. Hi4d~\cite{yin2023hi4d} focuses on human-human interactions, employing off-the-shelf implicit avatars~\cite{chen2021snarf} to model each subject individually. Like NeuralDome, these methods encounter obstacles in terms of deployment and generalization. 

In this work, we aim to explore the intricate, instance-level reconstructions of human and objects, along with the realistic contact surface deformations. As a testbed for our methods, we conducted experiments in the frequently encountered scenario of human-chair interactions. To our knowledge, COUCH \cite{zhang2022couch} and CHAIRS \cite{jiang2023chairs}, have investigate interactions between humans and chairs. However, both of these datasets focus on dynamic motions, which are captured through marker-based motion capture systems. They are unable to capture fine-grained geometry like casual suit and are not well suited for visual tasks like 3D reconstruction due to the markers attached to the human body. Therefore, we have collected a new dataset named Ins-Sit. Our dataset, in contrast, captures static poses, allowing us to focus on things like texture or geometry of people's clothes, and details when human interact with chair. These static scans and dynamic motion sequences are mutually complementary.

\subsection{Hand Object Interactions}
In the realm of hand-object interactions, significant strides have been made towards understanding and predicting hand object poses from images or videos. Numerous datasets have been proposed~\cite{brahmbhatt2020contactpose, dreher2019learning, fan2023arctic, hampali2020honnotate, hasson19_obman, jian2023affordpose, Liu_2022_CVPRhoi4d, GRAB:2020, XIE2023101178hmdo, YangCVPR2022OakInk, liu2024taco, ye2021h2o, Li_2023_ICCV_CHORD}, encompassing a wide variety of formats and scenarios, including images, videos, single-hand and bimanual hand, as well as interactions with rigid objects, articulated objects, and non-rigid objects, from single-view to multi-view perspectives. Based on the datasets introduced, numerous related studies have been conducted, exploring various aspects of hand-object interactions. Methods~\cite{hampali2020honnotate, YangCVPR2022OakInk, chao:cvpr2021, 6126483, wangyh}, which utilize multi-view setup to track human and object poses, achieve plausible results. 
There are also many works~\cite{ye2022hand, hasson19_obman, yang2021cpf, GraspingField:3DV:2020, chen2022alignsdf, chen2023gsdf, tse2022collaborative, hu2022physical, zhang2021single, zhao2022stability, hu2023learning} based on monocular setups, which are quite challenging due to the limited visual cues. In the majority of these works, there is an assumption that the object model is obtained beforehand, and they predict its 6D poses at test time. Although they can get plausible geometry of objects by this way, the deformations caused by contact compression are inherently overlooked. Some work~\cite{ye2022hand} relax this assumption, and predict the object's geometry based on visual clues and hand pose, yet the results remain limited. More related work to ours is HMDO~\cite{XIE2023101178hmdo}, which explored object deformation upon contact. However, their work still relies on the assumption that object template is given during model inference. Overall, current works in hand-object interactions more or less require the usage of hand templates~\cite{MANO:SIGGRAPHASIA:2017} or object templates. Our goal is to completely eliminate the dependency on templates, aiming to directly and separately reconstruct the geometry of both hand and object at test time, while restore the deformations caused by contact based on visual clues.  

In this work, to demonstrate the versatility of our instance-level reconstruction method, we also conduct validations on hand-object tasks. Despite the existence of numerous datasets, none of them can provide real textured meshes of authentic hand-object interactions. They either lack the geometry and texture of the hand (limited by parametric model representation) or lack the texture of the object, which made them unsuitable for 3D reconstruction tasks. To address this gap, we collect a new dataset, which consists of high-fidelity scans of hand-grasping objects. This significantly compensates for the current lack of high-precision meshes in hand-object interaction datasets. 

\begin{figure}[t]
  \centering
   \includegraphics[width=1.0\linewidth]{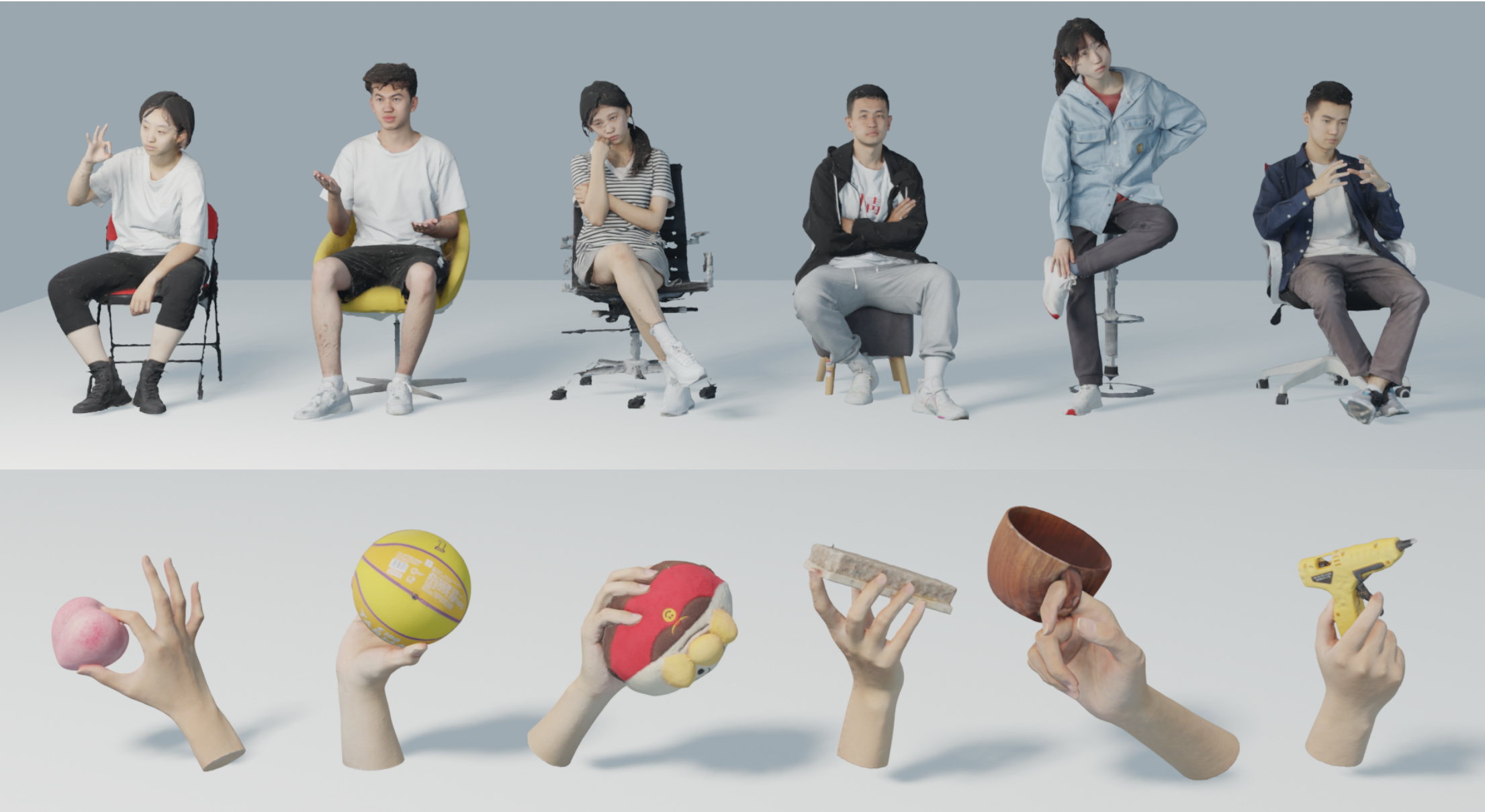}
   \caption{Examples of 3D scans from the Ins-Sit and Ins-Grasp dataset. It contains high-fidelity geometries and textures. Ins-sit captures a wide range of sitting postures and diverse clothing style, whereas Ins-Grasp includes a broader range of objects.}
   \label{fig:dataset}
\end{figure}

\subsection{Implicit function field}
The integration of deep neural networks with 3D representations, specifically through explicit and implicit expression, significantly enhances the ability to capture, shape, and understand complex 3D structures. Explicit representations, such as point clouds~\cite{qi2016pointnet, lin2018learning}, voxel grids~\cite{maturana2015voxnet, su2015multi}, and triangular meshes~\cite{wang2018pixel2mesh}, directly encode surface and volume data. However, the application of neural networks to explicit representations often encounters challenges related to scalability, memory efficiency, and the handling of complex topologies. Implicit representations, in contrast, offer flexible and compact alternative by defining 3D shapes as level sets of continuous functions~\cite{carr2001reconstruction, curless1996volumetric}, such as signed distance functions or occupancy fields. Inspired by many human reconstruction works~\cite{pifuSHNMKL19,zheng2020pamir, xiu2022icon, shao2022doublefield}, which focuses on human reconstruction from single or sparse view RGB inputs, in this work, we adopt the form of implicit functions to recover better geometry details.
\section{Dataset}
\label{sec:dataset}

Existing Human-Object Interaction (HOI) datasets are typically based on motion capture systems, where human/hand are represented through parametric models, and objects are characterized by pre-determined geometry and their 6D poses. However, the representation of parametric model and object templates inherently lacks the ability to capture fine geometry and soft deformations caused by contact. These limitations result in current datasets being unsuitable for 3D reconstruction tasks and hinders further investigation into idealized HOI modeling, which should integrate the complete 3D geometry of interaction instances and reasonable invisible contact areas.
To bridge this gap, we create two new datasets named Ins-Sit and Ins-Grasp for human-chair and hand-object interactions, respectively, as illustrated in Fig.~\ref{fig:dataset}. In this section, we elaborate on our data acquisition, data processing, and the details of the synthetic data augmentation utilized in complementary training~\ref{sec:semi training}.

\subsection{Real-scanned Data}
\label{sec:realdata}
Our capture system is composed of 128 synchronized DSLR cameras, uniformly positioned in a spherical rig configuration. The images obtained are processed using commercial software~\cite{realitycapture} to perform textured mesh reconstruction. Each resulting mesh is characterized by up to 500K triangular faces, ensuring a detailed representation of fine-grained geometry. We also captured accurate 3D mesh for each object, for chairs in the Ins-Sit dataset, our capture process mirrors that of the dataset itself, while for objects within the Ins-Grasp dataset, we employ a high-precision handheld 3D scanner. 

\begin{figure}
  \centering
   \includegraphics[width=1.0\linewidth]{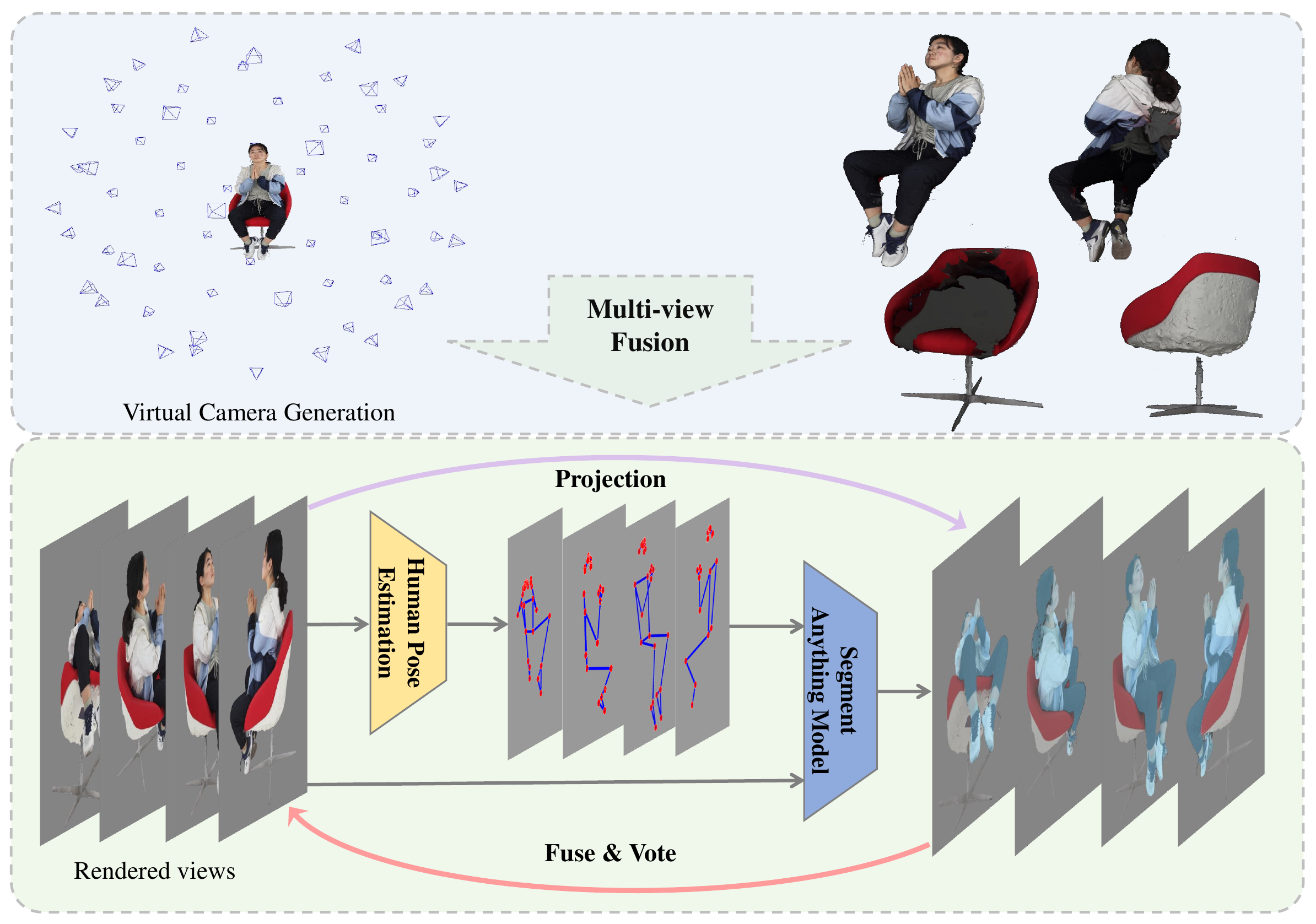}
   \caption{Pipeline of virtual multi-view fusion method for automatic 3D scan semantic segmentation.}
   \label{fig:seg}
\end{figure}

For human-chair interactions, the proposed Ins-Sit dataset comprises a total of 4700 scans. We invite 72 subjects (12 females and 60 males) as volunteers. Each subject seated on 2 of 11 distinct chair types and perform 60 unique poses. We fit SMPL-X~\cite{SMPL-X:2019} to human part in our scan through keypoints estimated from dense-view rendered images~\cite{openpose, mediapipe}. We also annotate the semantic label of each vertex in the scan by employing a combination of SAM~\cite{kirillov2023segany} and multi-view fusion~\cite{kundu2020virtual} techniques for evaluation and potential research purposes, as shown in Fig~\ref{fig:seg}.  Specifically, let $X_k \in \mathbb{R}^3$ represent the 3D position of the $k^{th}$ vertex. The pixel coordinates obtained by projecting the $k^{th}$ 3D point onto the virtual view $i \in I$ are denoted as $x_{k,i} \in \mathbb{R}^2$. For each view $i$, $K_i, R_i, t_i $ represents its intrinsic matrix, rotation matrix, translation vector, respectively.  Let $c_{k,i}$ denote the distance between the position of camera $i$ and the $k^{th}$ 3D point. The semantic label of the $k^{th}$ vertex $L_{3d}(k)$ can thus be defined as:
\begin{gather}
x_{k,i} = K_{i}(R_{i}X_{k} + t_{i}) \notag \\
c_{k,i} = \left\| X_{k} + R_{i}^{-1}t_{i} \right\|_2 \\
L_{3d}(k) = \text{vote}\left(\{ L_{2d}(k, i) \mid \|d(x_{k,i}) - c_{k,i}\| < \delta, i \in I \}\right) \notag
\end{gather}
where $ L_{2d}(k, i)$ corresponds to the 2D semantic label of the $k^{th}$ vertex in the $i^{th}$ view, $d(\cdot)$ maps pixel coordinates to their corresponding depth values from the rendered depth channel, and $\delta$ serves as a threshold for depth matching. In practice, we set the number of virtual cameras to 64, with these viewpoints uniformly distributed over a sphere of fixed radius. 2D keypoints detected by off-the-shelf detectors~\cite{openpose, mediapipe} are used as prompts for image segmentation~\cite{kirillov2023segany}, allowing for an automated process of semantic segmentation of 3D scans.

For hand-object interactions, the proposed Ins-Grasp dataset emphasizes more on the diversity of the objects due to the minimal variability in the appearance of human hands.
We collected data on 50 different objects, each involved in 10 distinct interactions. 
There are 20 volunteers participating in the data acquisition, contributing to a collection of 500 scans. We also fit MANO \cite{MANO:SIGGRAPHASIA:2017}, NIMBLE \cite{li2022nimble} and annotate the semantic label of each vertex with the same procedure as Ins-Sit.  

\subsection{Synthetic Data Augmentation}
\label{sec:data preparation}
The real-scanned dataset provides the interaction information as a single, connected mesh, which lacks individual ground truths.
Although we have semantically segmented the holistic mesh, it is still challenging to perform reasonable completion on meshes with extensive fragmented areas, as shown in Fig.~\ref{fig:seg}.
To support the learning of complete shape priors for our network, we propose to leverage individual scans of human, hand and objects to generate synthetic data. 

\begin{figure}
  \centering
   \includegraphics[width=1.0\linewidth]{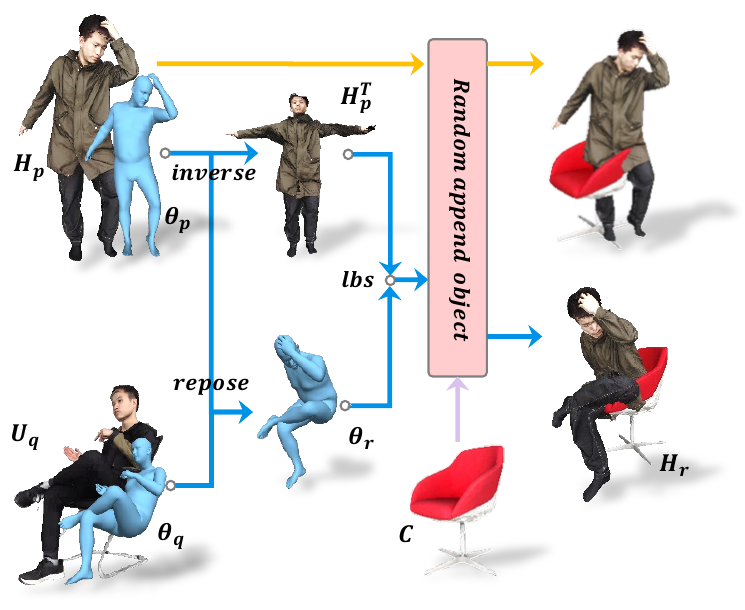}
   \caption{Examples and pipeline of our synthetic data generation process. For synthetic data of human-chair interactions, it comprises two distinct types: Syn\_s and Syn\_r.}
   \label{fig:synthesize}
\end{figure}

For human-chair interactions, we take a human scan $H_p$ from the existing high-quality human scans dataset THuman2.0~\cite{tao2021function4d} and couple it with an individual chair scan $C$ to imitate human-chair interactions, as shown in Fig.~\ref{fig:synthesize}.
As the subjects in THuman are mostly scanned under standing poses, to synthesize a more realistic seated posture data and reduce the gap between synthetic and real data, we further repose the standing human scans based on seated poses randomly chosen from our Ins-Sit.
To do so, we first apply the inverse LBS (Linear Blend Skinning) procedure to transform $H_p$ to T-pose as $H_p^T$. We then generate the reposed SMPL-X parameters $\theta_r$ by integrating the upper body pose of $\theta_p$ and the lower body pose of $\theta_q$. Next, we obtain the driven mesh of $H_p$ in pose $\theta_r$ as $H_r$, i.e.,
\begin{equation}
\left\{
\begin{aligned}
&\theta_r = concat\left(\left\{\theta^i_{p}| i \in upper body\right\}, \left\{\theta^j_{q}|j \in lower body\right\}\right) \\
&H_r = \mathcal{M}\left(\mathcal{M}^{-1}\left(H_p, \theta_p, W_{H_p}\right), \theta_r, W_{H_p}\right)
\end{aligned}
\right.
\end{equation}
where $\mathcal{M}$ denotes the LBS function, then $W_{H_p}$ is the LBS weight of $H_p$. For each vertex in $H_p$, we simply take the weight of the nearest SMPL-X~\cite{SMPL-X:2019} vertex in practice.
Based on the standing and reposed humans, we append them with pre-scanned chairs, following a set of rules. Specifically, we predefined ranges for random translation, rotation and scale, and then randomly sample values during data generation. For reposed sitting humans, given the reposed SMPL-X parameters and a fixed hip position index, we can approximate the hip's location and the chair seat height. This enables us to synthesize highly realistic sitting posture data. Notably, inspired by mixup~\cite{zhang2018mixup}, a common data augmentation technique in image classification and object detection tasks, this method increases the complexity of training data by blending images, thereby enabling the network to learn more robust capabilities. We allow for some interpenetration when arranging human and chair scans in 3D space. This help the network learn a more robust and strong individual shape priors. In summary, for human-chair interactions, our synthetic data comprises two types: standing and reposed sitting humans, denote as Syn\_s and Syn\_r, respectively. 

The similar synthetic strategy is also used to generate the augment data for hand-object interactions. Unlike human body reconstruction, where the diversity in clothing and poses can significantly influence the reconstruction results, the variation in hand appearance is minimal. Therefore, the processing of generating reposed data in human-chair interactions, is not necessary for hand-object interactions. Since MANO~\cite{MANO:SIGGRAPHASIA:2017} does not provide textures, we adopt NIMBLE~\cite{li2022nimble} as hand parametric representation. We first fit NIMBLE to collected scans to get the shape, pose parameters and combine them with randomly sampled appearance parameter to generate the realistic human hand. Subsequently, we combine the scanned objects with scanned objects following rules similar to those used in human-chair interactions, including rotation, translation and scale. 

\section{Method}\label{sec:methodology}

Given only sparse view images as input, our method Ins-HOI supports instance-level implicit reconstructions of two interacting objects, and recovers plausible contact surfaces in invisible areas. A major challenge in this problem is that, how can the network be trained to simultaneously predict the 3D occupancy fields of both instances in the absence of individual 3D ground truth. To tackle this, we initially define an instance-level neural surface field, using different branches within the same network to predict the occupancy of each instance. Subsequently, we introduce a complementary learning strategy, which leverages shape priors provided by synthetic data for mixed training with real data. 

\begin{figure*}[ht!]
  \centering
  \includegraphics[width=\linewidth]{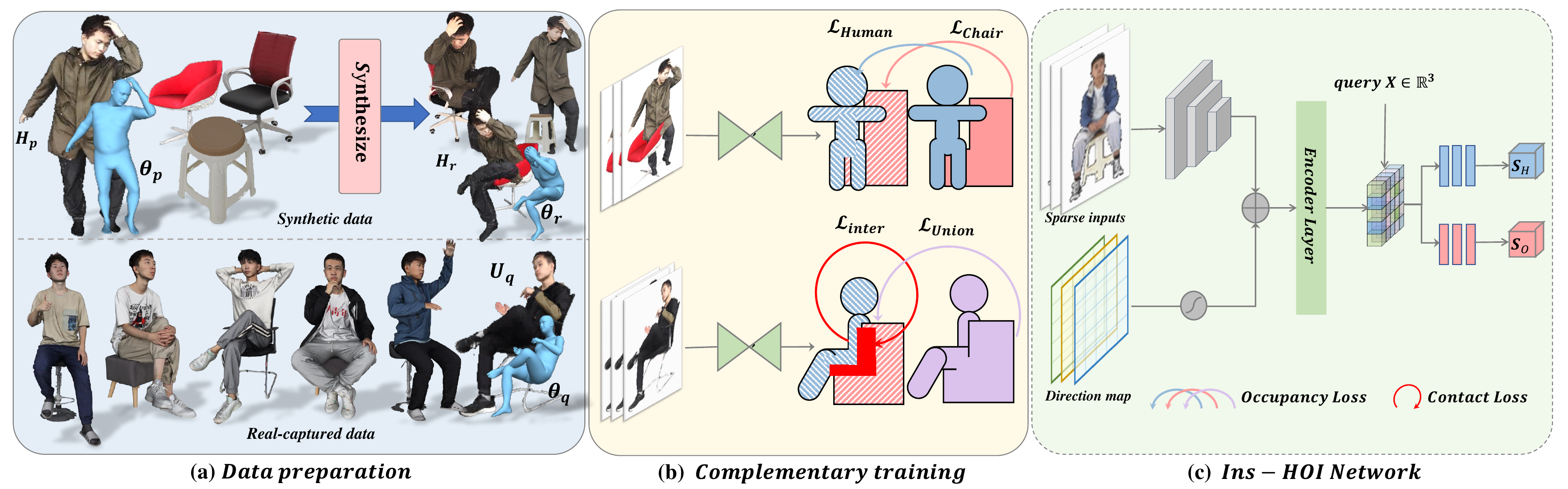}
  \caption{Overview of the method: (a) showcases the synthetic data augmentation process using THuman and Ins-Sit dataset to form a training dataset. (b) highlights how the training components provide unique guidance for complementary learning (blue and pink denote the human and chair meshes; purple and red indicate the union and intersection). (c) depicts our benchmark Ins-HOI, which given sparse view inputs to produce instance-level human-object recovery via an end-to-end approach.}
  \label{fig:method}
\end{figure*}

Ins-HOI is an unified and versatile method for both human-object and hand-object interactions. For clarity, we illustrate the pipeline of our method based on human-chair interactions, as shown in Fig.~\ref{fig:method}. In this section, we first introduce the proposed instance-level implicit functions (Sec.~\ref{sec:instance implicit function}), then the complementary training strategy (Sec.~\ref{sec:semi training}).

\subsection{Instance-level Neural Surface Field}
\label{sec:instance implicit function}

Our representation is built upon the pixel-aligned implicit functions similar to PIFu\cite{pifuSHNMKL19}, which defines a surface as a level set as follows:
\begin{equation}
f(F(x), \pi(X), z(X)) = s, s \in [0, 1],
\end{equation}
where $x = \pi(X)$ represents the projected 2D point in the image plane from the given 3D point $X$, $Z(X)$ denotes the depth value from $X$ to the camera, and $F(x)$ represents the pixel-aligned feature at $x$. 
Inspired by ~\cite{shao2022doublefield}, we further introduce a position embedding $\gamma$ of view direction $d(x)$ as extra input features.
The above pixel-aligned implicit functions only represent the target object as a single entity.
For instance-level reconstruction, we extend the implicit functions to support simultaneous representation of human/hand and objects.
Specifically, the proposed instance-level occupancy field has different outputs corresponding to human/hand and object as follows:
\begin{equation}
\left\{
\begin{aligned}
f_H(F(x), \gamma(d(x))) = s_H, s_H \in [0, 1] \\
f_O(F(x), \gamma(d(x))) = s_O, s_O \in [0, 1]
\end{aligned}
\right.
\end{equation}
Here, $s_H$ and $s_O$ refer to the occupancy fields of the human/hand and the object, respectively. 
The points inside human/hand and object mesh are represented by $H = \{X|f_H(X)=1\}$, and $O = \{X|f_H(X)=1\}$, respectively. 
Intuitively, $f$ predicts the continuous inside/outside probability field of a 3D model, from which an iso-surface can be easily extracted.

\begin{figure}
  \centering
  \includegraphics[width=0.9\linewidth]{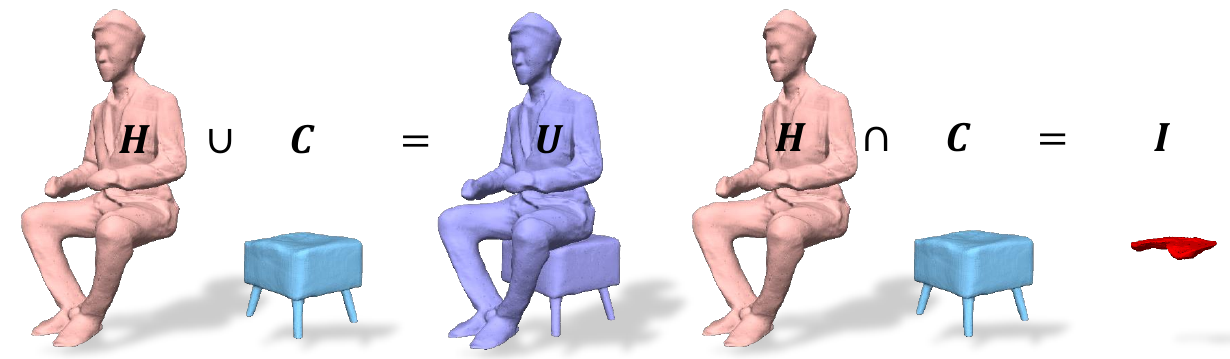}
   \caption{Example of Union and Intersection operation under our definition.}
   \label{fig:ui}
\end{figure}

Based on the proposed implicit human/hand and object representation, we can define their union and their intersection as follows:
\begin{align}
U &= \{X | f_{H}(X) = 1 \text{ or } f_{O}(X) = 1\}  \nonumber \\
  &= \{X | \max(f_{H}(X), f_{O}(X)) = 1\} \\
I &= \{X | f_{H}(X) = 1 \text{ and } f_{O}(X) = 1\} \nonumber \\
  &= \{X | f_{H}(X) \cdot f_{O}(X) = 1\}
\end{align}
An example of the instance-level occupancy field is shown in Fig.~\ref{fig:ui}, where their union and intersection are visualized for better illustration.

Based on our representation, we can directly supervise the union $U$ and penalize the intersection $I$ of two interacting objects during training. For inference, we reconstruct the meshes by densely sampling the occupancy field $s_H$ and $s_O$ respectively over the 3D space and extract the iso-surface at 0.5 level-set of a continuous occupancy field using the Marching Cube algorithm~\cite{lorensen1987marching}.

\subsection{Complementary Training}
\label{sec:semi training}
Based on the real-scanned and synthesized data described in Sec.~\ref{sec:dataset}, we further introduce a complementary training strategy for instance-level human-object recovery.
During the complementary training, our network leverages the two types of training data, namely the synthetic data \{Syn\_s, Syn\_r\}  and real-scanned data $\left\{U_q\right\}$.
Although synthetic data allows the network to learn a rough individual shape priors, it still faces two main issues. First, human reconstruction does not well generalize to variations in pose, making it challenging to transfer the priors of a standing posture to a seated posture. Second, synthetic data falls short in simulating and capturing intricate details of real interactions, offering only a rough visual resemblance. On the other hand, using only real-captured data means we can only supervise the union of the shapes, providing no guidance on instance information. During the process of mixed training, the two types of data play complementary roles, with each providing unique information, as shown in Fig.~\ref{fig:method}~(b).

The synthetic data comprises the information of individual human/hand $H_p$ and the object $C$, which offers strong guidance for the learning of human and object implicit functions $f_H(X)$ and $f_C(X)$:
\begin{equation}
    \mathcal{L}_i = \frac{1}{n} \sum_{i=1}^{n}({\left|f_H(X)-f^{*}_H(X)\right|^2} +  {\left|f_C(X)-f^{*}_C(X)\right|^2})
\end{equation}
where $n$ is the number of point samples, and $*$ represents the ground truth occupancy value. 

The real-scanned data is taken from the proposed Ins-Sit/Grasp scan dataset.
Due to the lack of instance-level ground truth, direct supervision of the individual human/hand and object is not feasible.
As the real-scanned data offers actual interaction information in the form of unified human-chair and hand-object meshes, we leverage such data by imposing supervision on the unioned implicit function and penalty on their intersection.
Specifically, the supervision on the union surface enhances the reconstruction of the interacting human/hand and objects as a whole, where the learning objective is defined based on $f_U(X) = \max(f_H(X), f_C(X))$:
\begin{equation}
    \mathcal{L}_u = \frac{1}{n} \sum_{i=1}^{n}{\left|\max(f_H(X), f_C(X))-f^{*}_U(X)\right|^2}
\end{equation}
In addition, we introduce the penalty on the intersection $I = H \cap C$ encourages an empty set, such that the predicted human/hand and object meshes are separated from each other:
{\small
\begin{equation}
    {
    \mathcal{L}_{in} = \frac{1}{n} \sum_{i=1}^{n}{\max(0, f_C(X)-0.5)^{1-\gamma} \cdot \max(0, f_H(X)-0.5)^{\gamma}}
    }
\end{equation}
}
Here $\gamma$ is introduced as a hyper-parameter to control the rigidity of the objects. 
For instance, when $\gamma$ is set to 1, the gradient $\frac{\partial \mathcal{L}_{in}}{\partial f_C}(X)|_{\gamma = 1} \equiv 0$, which means that only the human occupancy $f_H(X)$ is penalized to prevent intersection. Since the object's prior is learned from synthetic data, where the objects are in rest pose, it results in a more completed object without deformation.
Conversely, when $\gamma$ is set to 0, the gradient $\frac{\partial \mathcal{L}_{in}}{\partial f_H}(X)|_{\gamma = 1} \equiv 0$, which means only the object occupancy $f_C(X)$ is penalized, resulting in more intensive deformation of the objects.
In practice, we annotate the value of $\gamma$ for different types of chairs and objects in our dataset. 
This allows us to inject the knowledge of object materials and control the degree of deformation when a person is seated or grasping with hand, which will be further discussed in our experiments. 
In summary, the entire training objective can be written and minimized as:
\begin{equation}
    \mathcal{L}_{total} =  \mathcal{L}_{i} +  \mathcal{L}_{u} + \mathcal{L}_{in}
\end{equation}
\section{Experiments}\label{sec:experiments}

Ins-Sit and Ins-Grasp datasets serves multiple purposes, enabling 3D reconstruction, novel view synthesis, motion capture, motion generation, etc. In this paper, we benchmark our dataset on instance-level reconstruction tasks in this paper. We first introduce the dataset and evaluation metrics used for our tasks. Then we describe the selected baselines and provide comparisons against them. Furthermore, we conduct the ablation studies to analyze the effects of intersections and synthetic data. For more free-view and dynamic results, we encourage viewers visit our project page for more information.

\subsection{Datasets and Evaluation metrics}
\label{sec:eval metric}
\paragraph{Dataset.} We conduct experiments on our collected dataset, Ins-Sit and Ins-Grasp. 
\begin{itemize}
    \item \textbf{Human-Chair Interaction.} Ins-Sit consists of 4700 3D textured meshes from 72 subjects. We use 4000 for training and 700 for testing. Furthermore, we adopt two distinct data partitioning for evaluation of test data. In the first setting, for each subject, we select 5 poses to form the testset, with the remainder used for training. This results in \textbf{Within-subject} testset comprising 330 scans. To further evaluate the model's generalization ability to new subjects, we choose 66 subjects used for training and 6 subjects for testing. This results in \textbf{Cross-subject} testset comprising 407 scans.
    \item \textbf{Hand-Object Interaction.} Ins-Grasp consists of 500 scans, featuring 50 objects with 10 unique poses per object. Given that the appearance of hands is similar, the reconstruction results are not significantly affected by subject IDs. Therefore, we adopt a single data partition. For each object, we select 8 poses for training and 2 for testing, resulting in a testset that consists of 100 scans. 
\end{itemize}
\paragraph{Evaluation Metrics.} For evaluation, we take cues from well-regarded research on clothed human reconstruction~\cite{pifuSHNMKL19, zheng2020pamir, xiu2023econ}. Following them, we leverage Chamfer Distance and Point-to-Surface as our key metrics. 
\begin{itemize}
    \item \textbf{Chamfer Distance} Chamfer distance measures the average closest point distance between two sets of points. It serves as a symmetric metric, evaluating the similarity by calculating both the distance from each point in one set to its nearest point in the other set and vice versa.  
   \item \textbf{Point-to-Surface} The point-to-surface (P2S) distance calculates the average distance from each point in a predicted shape to the nearest surface point of a reference shape. This metric assesses the accuracy of the reconstructed geometry by measuring how close the points of the generated model are to the ground truth surface.
    \item \textbf{Selective Instance Metrics} As mentioned in Sec.~\ref{sec:realdata}, we annotate the semantic label of each vertex in the scan. To evaluate the reconstruction of the visible parts, we compute the one-directional Chamfer distance $CD(A, B) = \frac{1}{|A|} \sum_{a \in A} \min_{b \in B} \|a - b\|_2$ from the semantically segmented mesh $A$ to our instance-level reconstruction results $B$.  
\end{itemize}

\subsection{Implementation Detail}
\paragraph{Network architecture.} We train our network by synthetic rendering multi-view images from our scan datasets. In our paper, all experiments are conducted under 6 view settings, uniformly selected from a circle. We render the images by aligning subjects to the image center using a perspective camera model, which enables deployment in real-world settings, with an image resolution of 512$\times$512. As illustrated in Fig.~\ref{fig:method}(c), for image feature extraction, we adapt the same 2D image encoders, specifically the stacked hourglass~\cite{newell2016stacked} architecture, which is same to that used by the methods we compare against. Subsequently, we fuse multi-view pixel-aligned image features using a transformer encoder~\cite{vaswani2017transformer}. Finally, the implicit functions that responsible for each instance are implemented by multi-branch, multi-layer perceptron. 

\paragraph{Synthetic data generation.} For \textbf{human-chair interactions}, we employ two types of synthetic data as detailed in Sec.~\ref{sec:data preparation} and Fig.~\ref{fig:synthesize}, i.e., Syn\_s and Syn\_r. For Syn\_s, we iterated over each of the 500 human subjects in THuman2.0 dataset~\cite{tao2021function4d}, randomly selecting 5 out of 11 chairs for data synthesis. For Syn\_r, the process mirrors that of Syn\_s, but the poses of the lower body were randomly sampled from ours Ins-Sit dataset to perform reposing of human. This yield two synthetic datasets, Syn\_s and Syn\_r, each comprising 2500 scans. Our method requires synthetic data for the training process, unlike the comparative methods which do not. However, synthetic data also enhances the diversity of our training samples. Therefore, for a fair comparison, we included synthetic data in the training process when retraining our comparative methods.  For \textbf{hand-object interactions}, the Ins-Grasp dataset contains 50 objects. For each object, we randomly sample 10 hand poses from 500 available in the Ins-Grasp dataset. We augmented the pose and shape parameters of the NIMBLE~\cite{li2022nimble} model and randomly generated some textures to create hand models, which were then combined with objects. This yield a synthetic dataset containing 500 samples.

\begin{figure*}[ht!]
  \centering
  \includegraphics[width=1.0\linewidth]{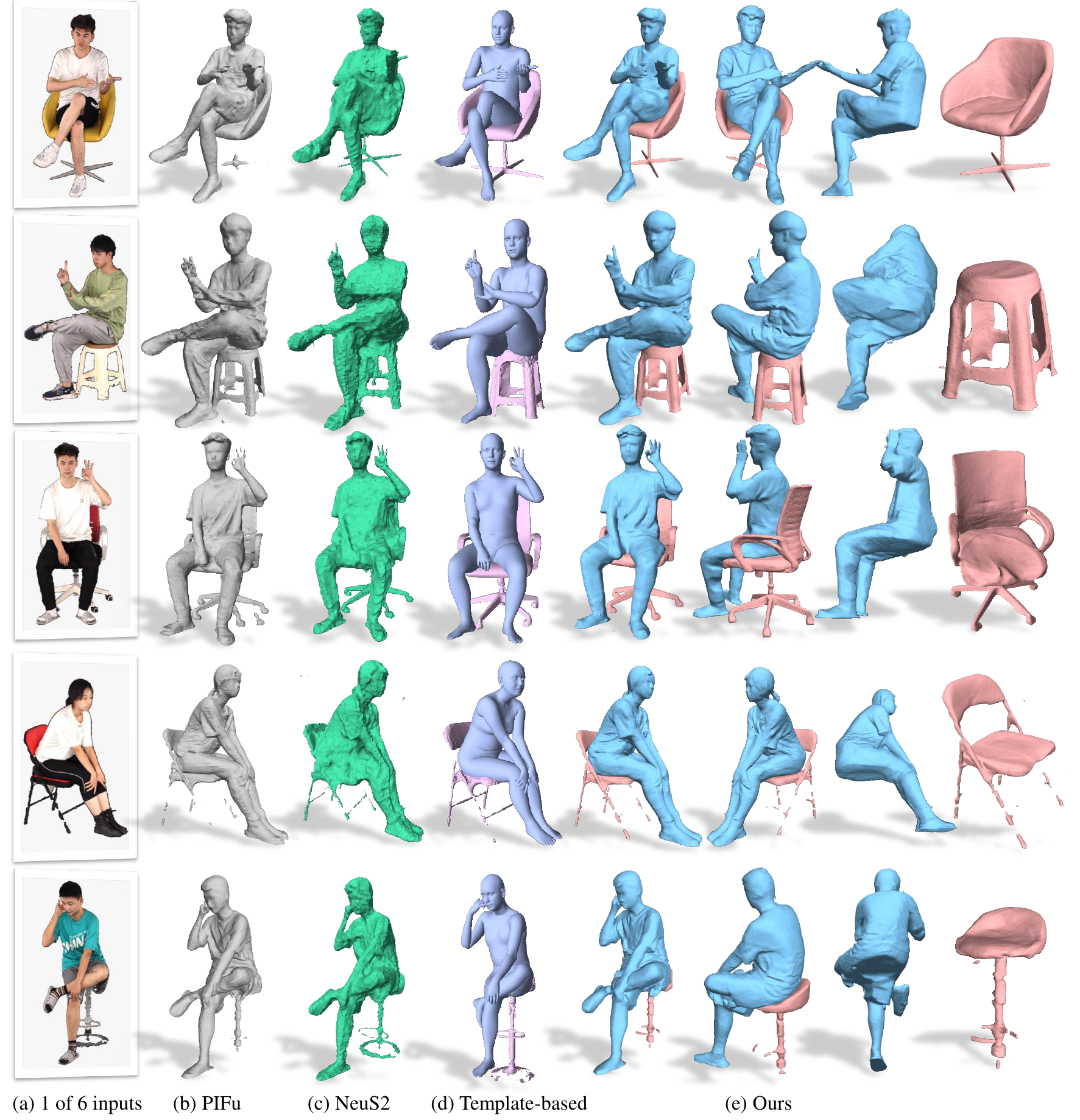}
  \caption{Qualitative comparison of geometric reconstruction results with PIFu \cite{pifuSHNMKL19}, NeuS2\cite{neus2} and template-based representations on Ins-Sit dataset.}
  \label{fig:results}
\end{figure*}

\paragraph{Network training. } We employ Adam~\cite{kingma2014adam} optimizer with an initial learning rate of 0.0001. Each subject is represented by 6000 randomly sampled points. Our training utilizes 4 GPUs, with a batch size of 4. The implementation is based on PyTorch~\cite{paszke2019pytorch}.  The $\gamma$ coefficient, as described in Sec~\ref{sec:semi training}, controls the degree of deformation for each object in the contact areas. For each interacting object, we assigned different $\gamma$ coefficients based on their material. For rigid objects such as teacups, we set the $\gamma$ coefficient to 1. For flexible objects like basketballs, the coefficient was adjusted to 0.75. For very soft items, such as plush toys and sofa chairs, we assigned a $\gamma$ value of 0.5.

\subsection{Benchmark on Geometry Reconstruction}
As our method is the first to enable instance-level human object interactions recovery with intricate geometry details, and considering we have collected two new datasets for this task, there are no existing baselines available for direct comparison. PIFu~\cite{pifuSHNMKL19} first introduce to utilize pixel-aligned implicit functions for human reconstructions and show its generalization capability. 
Recent years, NeRF~\cite{mildenhall2020nerf} has attracted considerable interest from researchers due to its high-quality results in the tasks of novel view synthesis but it cannot extract high-quality and the training process is slow. NeuS2~\cite{neus2} represents the 3D surface as a Signed Distance Function for high-quality geometry and supports fast training. Therefore, we select PIFu~\cite{pifuSHNMKL19} and NeuS2~\cite{neus2} as comparative methods in our paper, which include both implicit neural surface field based approach and neural radiance field based method. To more intuitively demonstrate the differences between our method and the previous template-based methods, we also present a comparison of using templates to model human-object interactions.

\begin{table*}[ht!]
    \caption{Quantitative comparison of 3D reconstruction methods on the Ins-Sit dataset. We evaluates the performance of each method using Chamfer distance and Point-to-Surface (P2S) metrics (cm) and their capability at the instance level. we also include the One-Dimensional Chamfer Distance (1D-CD) between the reconstructed human (H), chair (C) instances against the semantically segmented mesh of scan in visible parts.}
  \centering
    \begin{tabular}{l|cc|cc|c|cc|cc}
      \hline
      Methods & \multicolumn{2}{c|}{Cross-subject} & \multicolumn{2}{c|}{Within-subject} & Instance Level & \multicolumn{2}{c|}{Cross-subject} & \multicolumn{2}{c}{Within-subject}  \\
       (6 views) & Chamfer $\downarrow$  & P2S $\downarrow$  & Chamfer $\downarrow$  & P2S $\downarrow$  & &  1D-CD (H) & 1D-CD (C)  & 1D-CD (H) & 1D-CD (C) \\
      \hline
      PIFU~\cite{pifuSHNMKL19}& 0.5280 & 0.5660 & 0.5194 & 0.5265 & × & - & - & - & - \\  
      NeuS2~\cite{neus2} & 1.2193 & 1.4757 & 1.2498 & 1.5007 & × & - & - & - & -\\
      Ours & 0.4828 & 0.4743 & 0.4736  & 0.4683  & $\checkmark$ & 0.5371 & 0.6089 & 0.5302 & 0.6079\\  
      \hline
    \end{tabular}
  
  \label{tab:benchmark}
\end{table*}

In contrast to the instance-level reconstruction introduced in our paper, comparative methods typically yield a single connected mesh, which we refer to as holistic reconstruction. These methods can be directly trained and evaluated on our dataset to assess their reconstruction results. For instance-level reconstruction, our approach faces the challenge of lacking instance-level ground truth data for both training and evaluation. We address the training challenge with our proposed method. For evaluation, we validate our method in three aspects. First, we merge two interacting subjects as a whole and evaluate the holistic reconstruction results. Second, we evaluate the degree of interpenetration between the two interacting instances; the lower the interpenetration, the better the instances are separated at the contact area. Third, given the instance segmentation label on captured data, we assess the reconstruction results of each instance in their visible parts. Through these three quantitative metrics, combined with qualitative results, we believe that our instance-level reconstruction results can be well evaluated.

\textbf{Human-Chair Interaction. } We conduct qualitative comparison in Fig.~\ref{fig:results}. In the figure, we present the results from Ins-Sit test set. PIFu~\cite{pifuSHNMKL19} and ours Ins-HOI are generalizable approaches to 3D reconstruction, whereas NeuS2~\cite{neus2} is scene-specific. Our reconstruction results are visually superior to those of PIFu, NeuS2. Additionally, our method supports instance-level reconstruction. In the visible parts, our method not only achieve high-quality geometry reconstruction but also effectively separates each instance distinctly. Moreover, we achieve plausible and realistic reconstruction of invisible contact surfaces with non-rigid deformations caused by compression due to sitting. In Fig.~\ref{fig:results}(c), we show the template-based representation of Ins-Sit by fitting the human body with SMPL-X~\cite{SMPL-X:2019} and aligning a pre-scanned chair with the chair part using ICP~\cite{besl1992methodicp} registration. Template-based representation is limited by the inherent expressive capacity of parametric model, which lacks geometric details such as clothing and muscle deformations caused by compression. Besides, for the chair part, especially articulated components like casters, cannot be well-aligned, and the non-rigid deformations of the chair surface due to the compression cannot be modeled accurately. 

We also quantitatively compare our methods with three metrics in Tab~\ref{tab:benchmark}. Similar to PIFu~\cite{pifuSHNMKL19}, we use point-to-surface and Chamfer distance to evaluate the holistic reconstruction. Besides, we also assess the reconstruction quality of each instance's visible parts by evaluating the one-directional Chamfer distance based on our semantically segmented mesh, as described in Sec.~\ref{sec:eval metric}.  The quantitative comparison shows that ours Ins-HOI achieves comparable performance despite the absence of direct supervision. Ins-HOI surpasses PIFu and NeuS2 across all metrics, and uniquely supports instance-level reconstruction. The results in Tab.~\ref{tab:benchmark} demonstrate that our reconstruction achieve plausible results both in terms of holistic geometry and on the instance visible parts. In Tab.~\ref{tab:intersection}, we further evaluated the extent of intersection between two instances. As indicated in the table, the Intersection over Union~(IoU) between the two instance meshes and the volume of intersection are extremely low. This proves that our method effectively separate two closely interacting instances within the 3D spaces, indicating that the interacting instances share no penetration and the contact regions are closely approximates the actual ground truth.

\begin{figure}[ht!]
  \centering
  \includegraphics[width=1.0\linewidth]{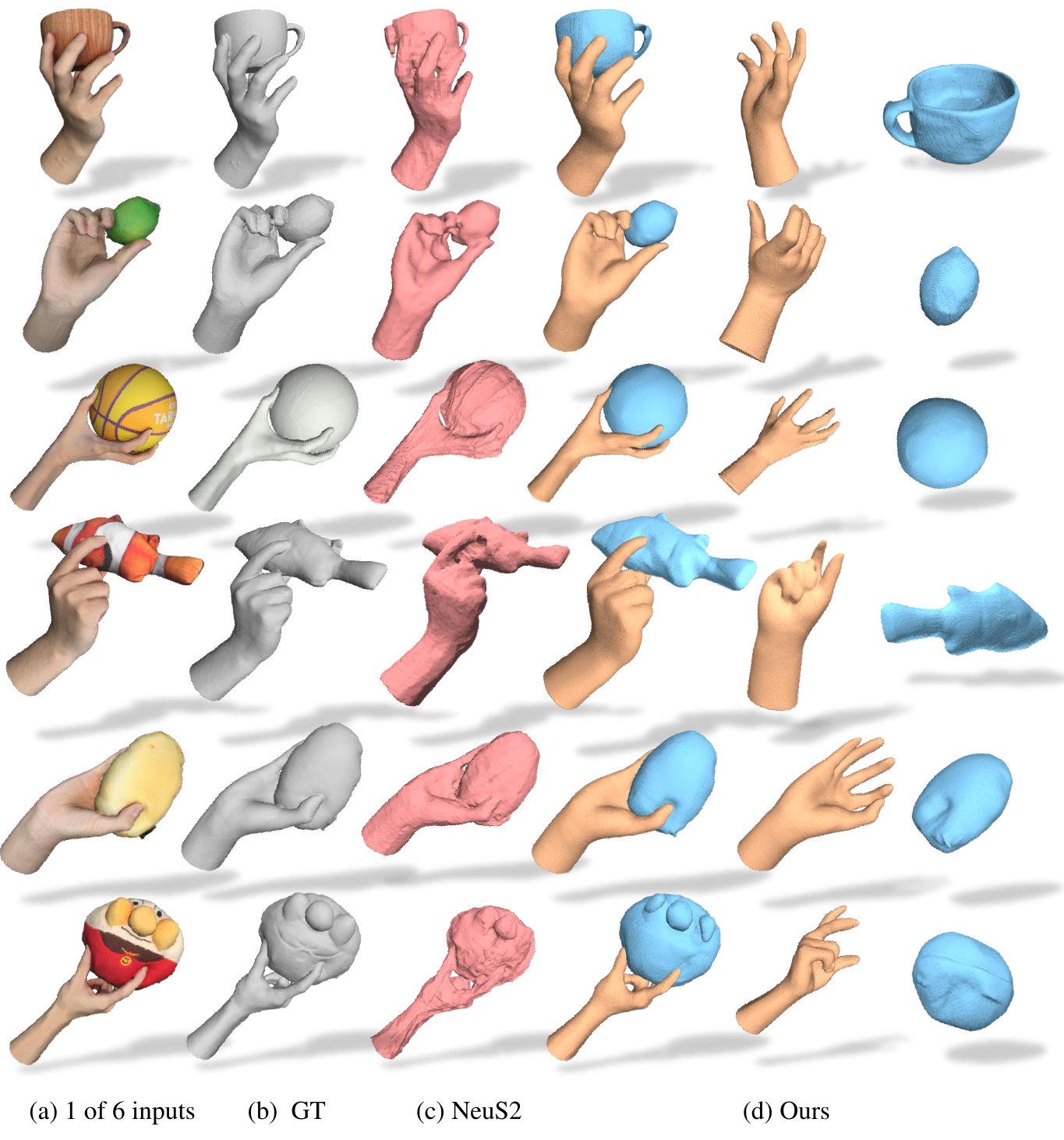}
  \caption{Qualitative comparison of geometric reconstruction results between NeuS2\cite{neus2} and ours on Ins-Grasp dataset.}
  \label{fig:hand results}
\end{figure}

\textbf{Hand-Object Interaction. }
Our proposed method is versatile, applicable not only to human-chair interaction scenarios but also to hand-object interactions, enabling instance-level reconstructions with soft deformation on the contact surfaces. Qualitative comparison is shown in Fig.~\ref{fig:hand results}, we compare our method with NeuS2~\cite{neus2}. We trained NeuS2 under default settings, for each scan in test set, we train 30000 steps to ensure full convergence. Our method is superior to NeuS2 in terms of the reconstructed geometry and support instance-level reconstruction. Notably, compared to the Ins-Sit dataset, Ins-Grasp dataset is significantly smaller in scale with 500 scans but encompass a wider variety of objects, including 50 types. For each type of object, the training set comprises only 8 scans. Despite the small size of the training set, the results on test set are plausible, demonstrating the generalization ability of our method. Another noteworthy point is the soft deformation, as illustrated in Fig~\ref{fig:hand results}, where the first three rows show rigid objects, such as cups, lemon, and basketball. Due to the different $\gamma$ coefficients set for objects of varying materials, there is no deformation in the contact areas between the hand and the objects. In contrast, the last three rows feature plush toys, which are made of softer material. The deformation caused by contact compression have also been accurately recovered, and these deformations are consistent with the visual clues provided by the input images. Qualitatively, we benchmark the results of geometric reconstruction on the Ins-Grasp dataset, which is same as Ins-Sit, as shown in Tab.~\ref{tab:benchmark_hand}.  For each of the 100 scans in the test set, we trained a NeuS2 model and report its average Chamfer distance and P2S metrics. As observed, our method significantly outperforms NeuS2, and also shows low interpenetration between the two instances. These indicate that our method is also well-suited for the task of hand-object interaction, yielding plausible results. 

\begin{table}
    \caption{Quantitative comparison of 3D reconstruction methods on the Ins-Grasp dataset. We evaluates the performance of each method using Chamfer distance and Point-to-Surface (P2S) metrics (cm) and report the degree of intersection between two reconstructed instances.}
  \centering
    \begin{tabular}{l|cc|cc}
      \hline
      Methods & \multicolumn{2}{c|}{Geometry} &   \multicolumn{2}{c}{Intersection} \\
       (6 views) & Chamfer $\downarrow$  & P2S $\downarrow$  & IoU($\%$) $\downarrow$ & Volume($m^3$) $\downarrow$ \\
      \hline
      NeuS2~\cite{neus2} & 0.2706 & 0.3907   & - & - \\
      Ours & 0.1535 & 0.1719   & $ 0.026\%$ & $3.84e^{-7}$\\
      \hline
    \end{tabular}
  
  \label{tab:benchmark_hand}
\end{table}

\subsection{Ablation Study}
\label{subsec:ablation}
\begin{figure}
  \centering
  \includegraphics[width=1.0\linewidth]{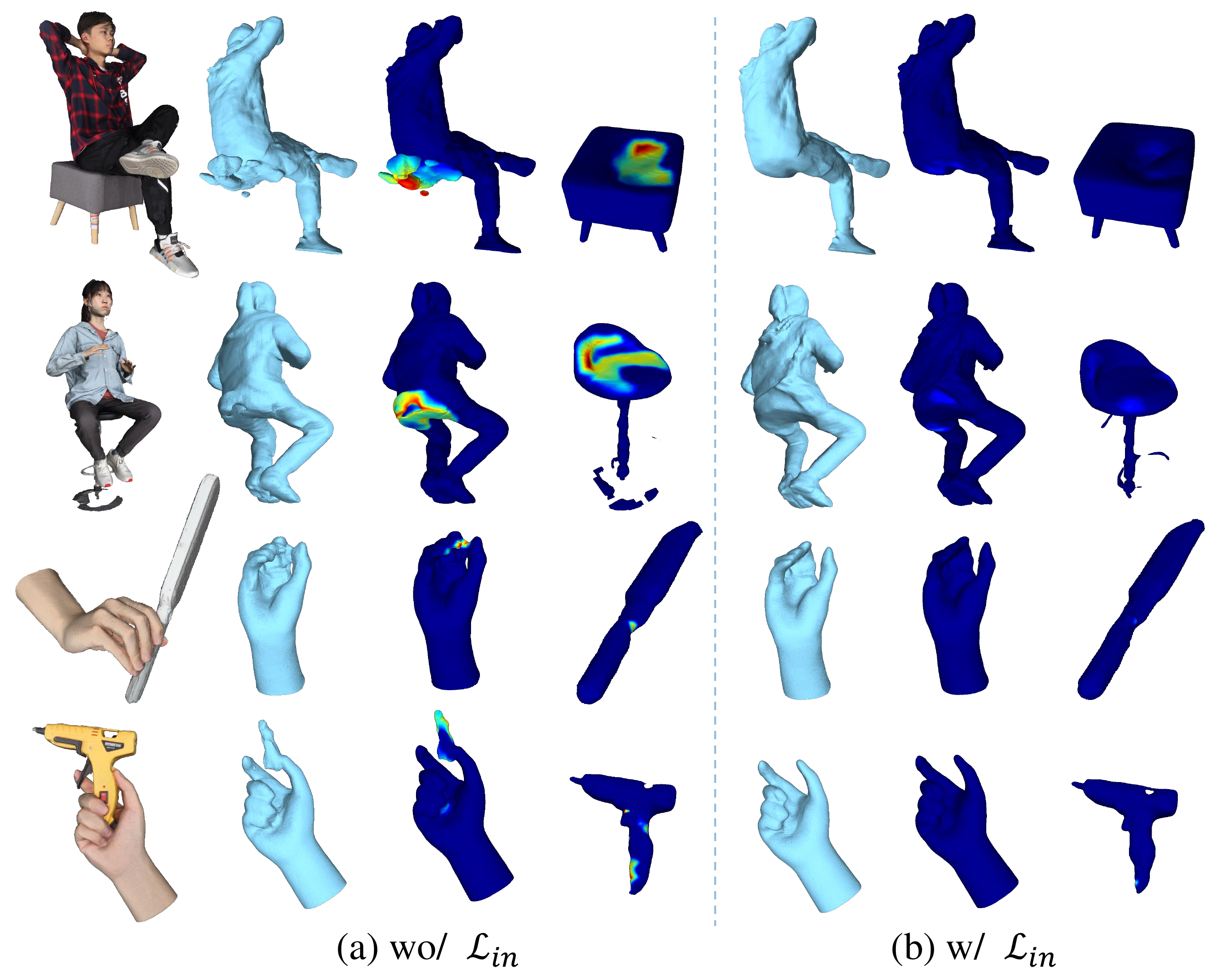}
   \caption{Qualitative evaluation of intersection constraint $\mathcal{L}_{in}$. We demonstrate the geometry and intersection heatmap in w/ and w/o $\mathcal{L}_{in}$ in (a) and (b), respectively.}
   \label{fig:ab_contact}
\end{figure}

\paragraph{Intersection constraint $\mathcal{L}_{in}$. } 
Despite the assistance of synthetic data in learning strong instance priors, interaction regions are often occluded, making the reconstruction of invisible areas challenging. Without addressing the potential overlap of the two occupancy fields, the human/hand is likely to intersect with the object in contact areas.
To investigate the effects of the intersection constraint $\mathcal{L}_{in}$ we proposed in preventing interpenetration, we conduct ablation experiments $w/$ and $w/o$ the intersection constraint. Quantitatively, in Tab.~\ref{tab:intersection}, we quantified the extent of interpenetration by measuring the Intersection over Union (IoU) and the volume of the overlapped regions on the Ins-Sit dataset. This demonstrated that with $\mathcal{L}_{in}$, the extent of of intersection is significantly reduced. Tab.~\ref{tab:benchmark_hand} also shows the quantitative results of intersection on the Ins-Grasp dataset, which is remarkably low. 
We also present qualitative results for a more intuitive description in Fig.~\ref{fig:ab_contact}, utilizing a heatmap visualization to illustrate mesh intersections. This heatmap encodes interpenetration depth, with warmer colors indicating areas of more significant overlap. Overall, it is clearly evident from our results that, with the contact constraints, the degree of mesh intersection is significantly reduced to minimal. 

\begin{table}[h]
    \caption{Quantitative evaluation of intersection degree w/o and w/
 $\mathcal{L}_{in}$ on Ins-Sit dataset.}
    \centering
    \resizebox{\columnwidth}{!}{
    \begin{tabular}{l|cc|cc}
         & \multicolumn{2}{c|}{Cross-subject} & \multicolumn{2}{c}{Within-subject} \\
         & IoU($\%$) $\downarrow$ & Volume($m^3$) $\downarrow$ & IoU($\%$) $\downarrow$ & Volume($m^3$) $\downarrow$ \\ 
         \hline
          w/o $\mathcal{L}_{in}$ & $8.1\%$ & $8.59e^{-3}$ & $7.65\%$ & $8.16e^{-3}$  \\
          w/ $\mathcal{L}_{in}$ & $0.058\%$ & $5.86e^{-5}$  & $0.055\%$ & $5.56e^{-5}$  \\
    \end{tabular}
    }
    \label{tab:intersection}
\end{table}

\paragraph{Prior intensity $\gamma$. }
To better reconstruct the inner contact surfaces, drawing from both observed image features and intrinsic properties of object material, we introduce the hyper-parameter $\gamma$ to guide network learning the extent of deformation. 
$\gamma$ is designed to modulate the network's penalization gradient across the occupancy fields of interacting objects, directly affecting the degree of object deformation, as detailed in Sec~\ref{sec:semi training}. The introduction of the $\gamma$ coefficient makes the deformation of contact surfaces controllable, with a higher setting applied for rigid materials and a lower one for softer materials. 

\begin{figure}[h]
  \centering
\includegraphics[width=1.0\linewidth]{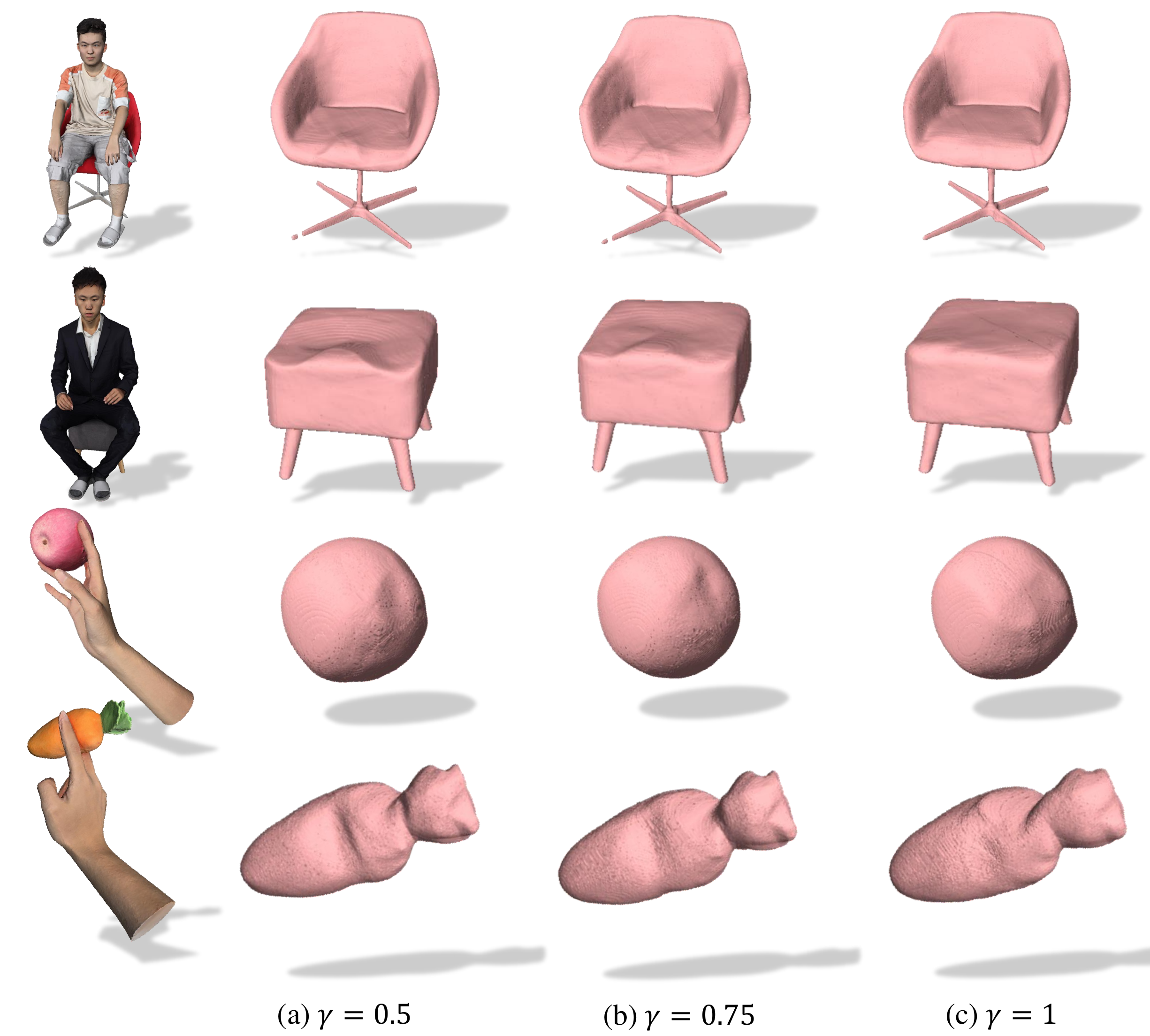}
   \caption{Ablation study of $\gamma$ coefficient. Variation of $\gamma$ control the penalization degree of human/hand and chair/object occupancy, thereby influence the extent of surface deformation.}
   \label{fig:ab_gamma}
\end{figure}

The varying effects of different $\gamma$ settings are shown in Fig.~\ref{fig:ab_gamma}. With $\gamma$ set to 0.5, an equitable balance in the penalization of the occupancy fields is applied to both chairs/objects and humans/hands. This means the deformability of both interacting entities is equalized, resulting in noticeable deformation of the chairs/objects.
When $\gamma$ is set to 0.75, the level of penalization applied to humans/hands is greater than that for objects, thereby reducing the deformation of the object compared to when $\gamma$ is at 0.5. This adjustment leads to less noticeable deformation in the object. When $\gamma$ is set to 1, the penalization within the interaction area's occupancy field is exclusively applied to the human/hand part. Consequently, the shape of the object is largely derived from the shape prior learned through synthetic data, specifically the object in its rest pose, resulting in smooth, undistorted and flat surfaces on chair/object. During the training process, we assign different $\gamma$ coefficient based on the material properties of the interacting objects to ensure realism and the optimal reconstruction results. 

\begin{figure}[h]
  \centering
  \includegraphics[width=1.0 \linewidth]{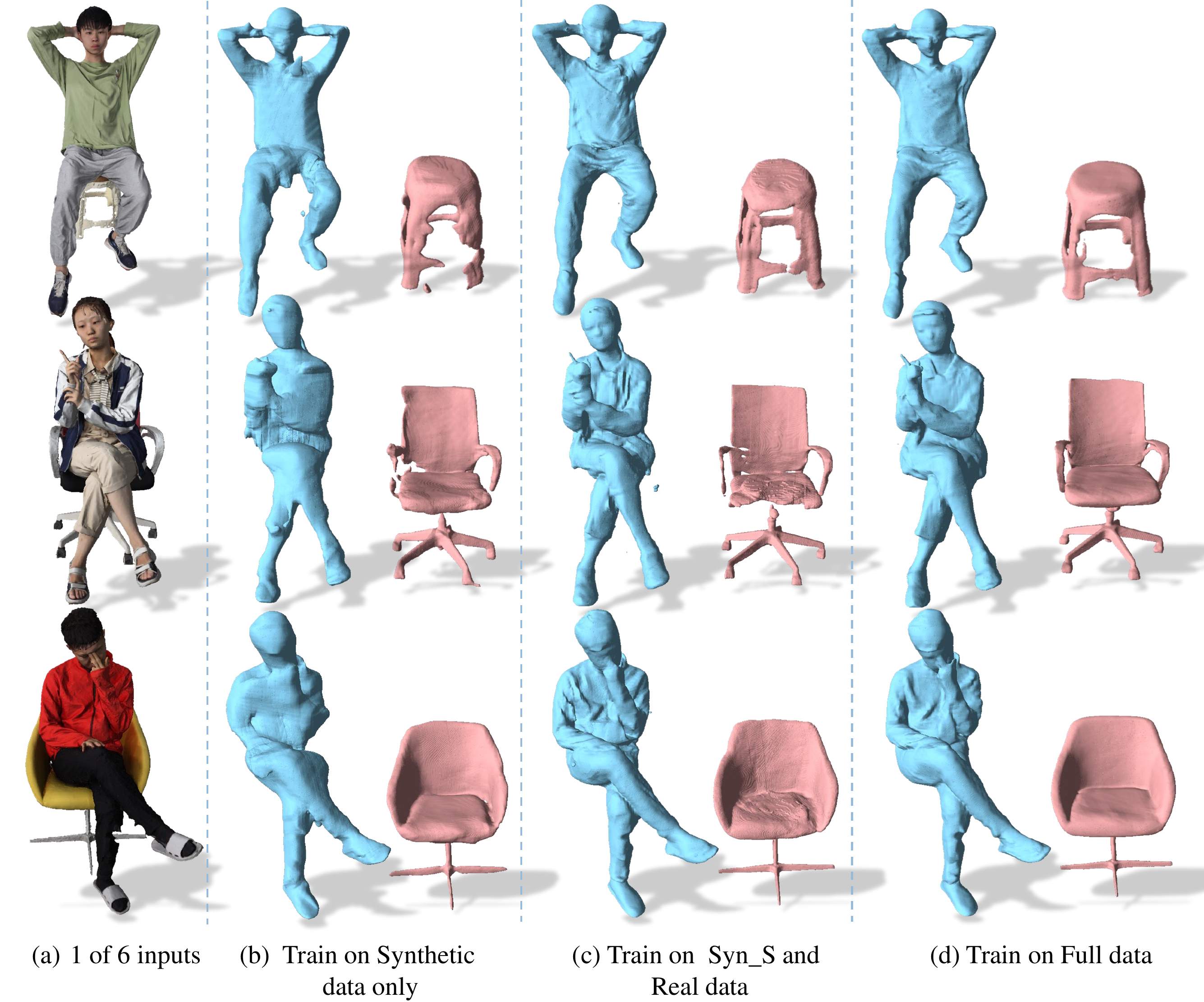}
   \caption{Ablation study on the contribution of each dataset types:(b)-(c) show reconstructions without real-captured data and reposed data, respectively. (d) includes all dataset types, leading to the most comprehensive result.}
   \label{fig:ab_dataset}
\end{figure}

\paragraph{Synthetic data augmentation. }
As illustrated in Fig.~\ref{fig:method}, our complementary learning approach is facilitated by two types of data: synthetic and real-captured. Synthetic data, composed of human/hand and object combinations in 3D space, providing ground truth of instance shapes, guiding the network in learning individual shape priors. Conversely, real-captured data ensures the integration of two instances seamlessly close to real scans. Both data types play distinct and indispensable roles during network training. Our ablation study highlights the distinct contribution of these data types. 

For human-chair interactions, the synthetic data is subdivided into two subcategories, as detailed in Sec.~\ref{sec:data preparation} and Fig.~\ref{fig:synthesize}. One includes standing humans with chairs, denoted as Syn\_S and the other comprises reposed humans and chairs, denoted as Syn\_R. We show the contribution of each data category in Fig.~\ref{fig:ab_dataset}.
As demonstrated in Fig.~\ref{fig:ab_dataset}(a), without real-captured data, the network can only learn coarse shape priors and is unable to learn detailed geometry of human in sitting pose and constraints of intersection.  Fig.~\ref{fig:ab_dataset}(b) shows that without repose synthetic data Syn\_R, the network struggles to learn a good prior for chairs due to a visual discrepancy between standing synthetic data and real data.  Overall, Syn\_S is used for the direct supervision of both human and chair occupancy, providing coarse shape priors. Syn\_R compensates for the visual discrepancy between the occlusion relationships of standing humans with chairs and those of sitting postures with chairs. However, during the repose process, the human mesh becomes non-watertight, making it unsuitable for network training. For Syn\_R, we only supervise the chair part, allowing this data help us learn strong priors of chairs in rest pose. The real captured dataset plays a crucial role in learning the fine geometry of seated poses. Although this dataset lacks instance-level ground truth, the geometry of seated poses can still be learned through the supervision of the union, based on the priors learned from the synthetic data.

\begin{figure}
  \centering
  \includegraphics[width=1.0 \linewidth]{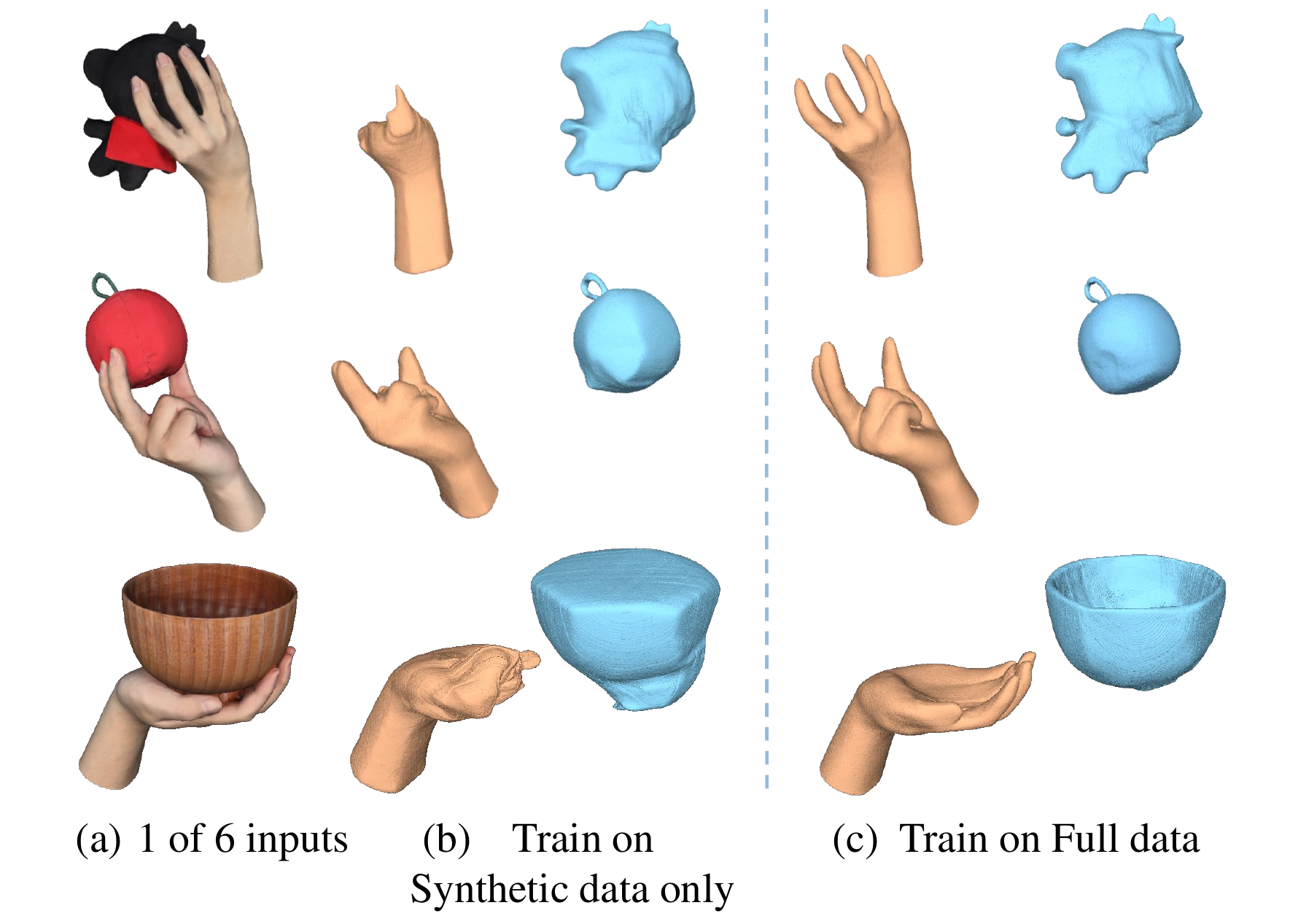}
   \caption{Ablation study on the contribution of each dataset types: (b) show reconstructions without real-captured data. (c) includes all dataset types, leading to more comprehensive result.}
   \label{fig:ab_dataset_hand}
\end{figure}

For hand-object interactions, the variation in the appearance of hands is not as significant as that of clothed humans, meaning the occlusion relationships between different people's hands and objects do not substantially impact the reconstruction of the object. Therefore, unlike human-chair interactions, we do not use reposed synthetic data. As shown in Fig.~\ref{fig:ab_dataset_hand}, when we solely train on synthetic data, the results are notably coarse, indicating that synthetic data serves to guide the network in learning the initial individual shapes. The refined geometry is derived from real-captured data. However, without synthetic data, our network would be incapable of training. Overall, synthetic data and real-captured data beneficially complement each other, enabling us to achieve better instance-level reconstruction results.

\subsection{Fine-tune Experiments}
\begin{figure}
  \centering
\includegraphics[width=1.0\linewidth]{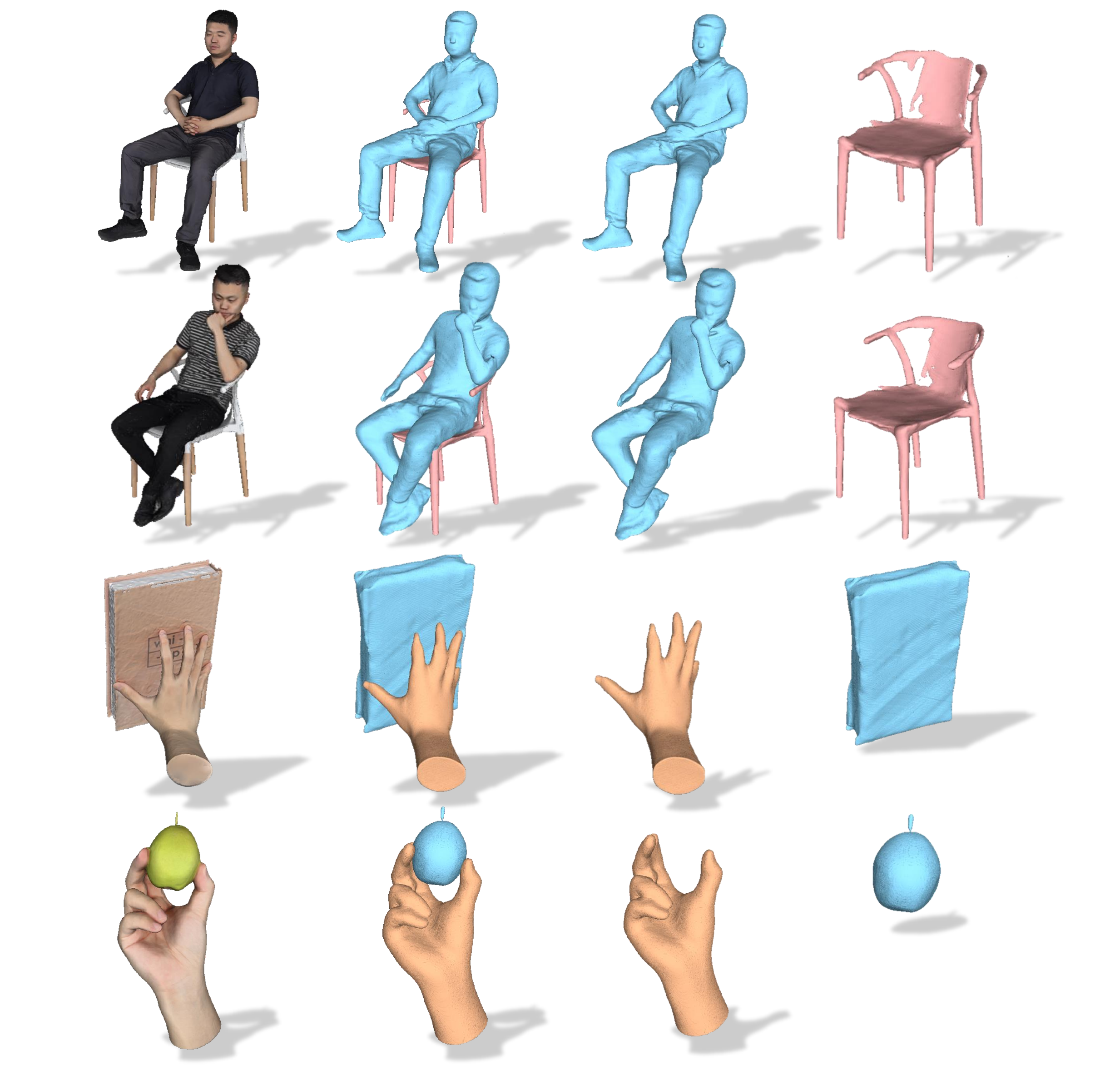}
   \caption{Experiments of fine-tuning on novel object, demonstrating that our model's ability to generalize to novel objects with limited amount of synthetic data for fine-tuning.}
   \label{fig:ab_finetune}
\end{figure}

The majority of research on human-object interactions relies on the assumption that object is given, which makes it challenging to generalize to new objects not included in the datasets. Many works on human reconstruction~\cite{pifuSHNMKL19, zheng2020pamir, xiu2023econ} have shown commendable generalizability. Our method Ins-HOI, trained on a large-scale dataset, can also well generalize to unseen human/hand. To validate the generalization ability of object aspect, we conducted a fine-tune experiments. When given a new type of chair/object, we synthesize a batch of data and perform fine-tuning on our pretrained model using synthetic data, and finally achieve plausible results. Specifically, for human-chair interactions, we synthesize 500 scans by integrating chair models with scans from THuman~\cite{tao2021function4d}, including both standing and resposed sitting poses. For hand-object interactions, we randomly choose 200 hand grasping poses from our Ins-Sit dataset. Utilizing the parametric NIMBLE model~\cite{li2022nimble}, we generate textured hand meshes and subsequently combine them with novel object. Approximately half an hour of fine-tuning, spanning 3-5 epochs, is sufficient to achieve plausible results, as shown in Fig.~\ref{fig:ab_finetune}.

\subsection{Experiments on Well-Established Datasets}
\begin{figure}
  \centering
\includegraphics[width=1.0\linewidth]{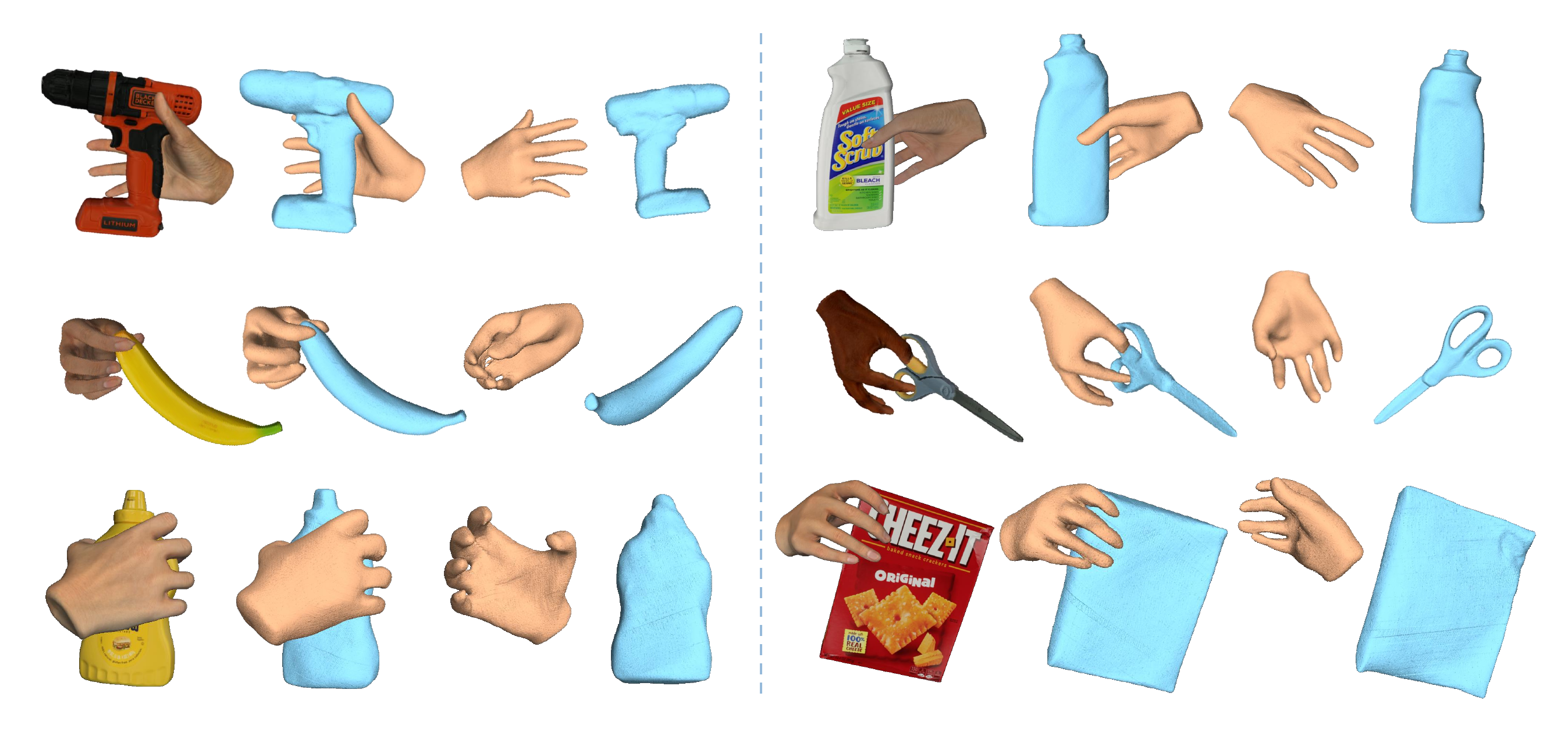}
   \caption{Qualitative results of the HO-3D dataset under our split test set configuration.}
   \label{fig:ho3d}
\end{figure}

In addition to conducting experiments on our dataset, we further validate our method on public available datasets. Due to the lack of high-fidelity, textured scans for human/hand-object interactions in existing datasets, we further processed the HO-3D~\cite{hampali2020honnotate} dataset to fit our training. The HO-3D dataset provides 3D poses of the hand and objects, with objects sourced from the YCB dataset~\cite{xiang2018posecnn} and hand represented by MANO~\cite{MANO:SIGGRAPHASIA:2017} hand model. However, the MANO model lacks hand textures, we fit MANO to another parametric model, NIMBLE~\cite{li2022nimble}, which supports generation of hand textures. For each sequence in the HO-3D dataset, we sample 50 frames to compose our training set. Of these, 30 frames are used to generate synthetic data with instance-level ground truth, incorporating random augmentations as described in Sec.~\ref{sec:data preparation}. The remaining 20 frames are used to merge two instances into a single connected mesh, which is in the same modality as the real-captured data in Ins-Grasp. Since the models in the YCB dataset are rigid objects (whose surfaces do not deform under contact pressure), we assign a gamma coefficient of 1. 
We showcase the qualitative results in Fig.~\ref{fig:ho3d}. The variety of objects and hand poses have been reconstructed individually by our model, and the geometric quality on HO-3D is as impressive as on the Ins-Grasp dataset. Additionally, it is important to note that the original HO-3D dataset exists numerous interpenetration at the area of interaction, due to the inherent limitations of parametric model representations. Nonetheless, our Ins-HOI, incorporated by contact constraints, have effectively reduced the cases of mesh penetration. 

\subsection{Results on Real World Multi-view Data}
\begin{figure*}
  \centering
\includegraphics[width=1.0\linewidth]{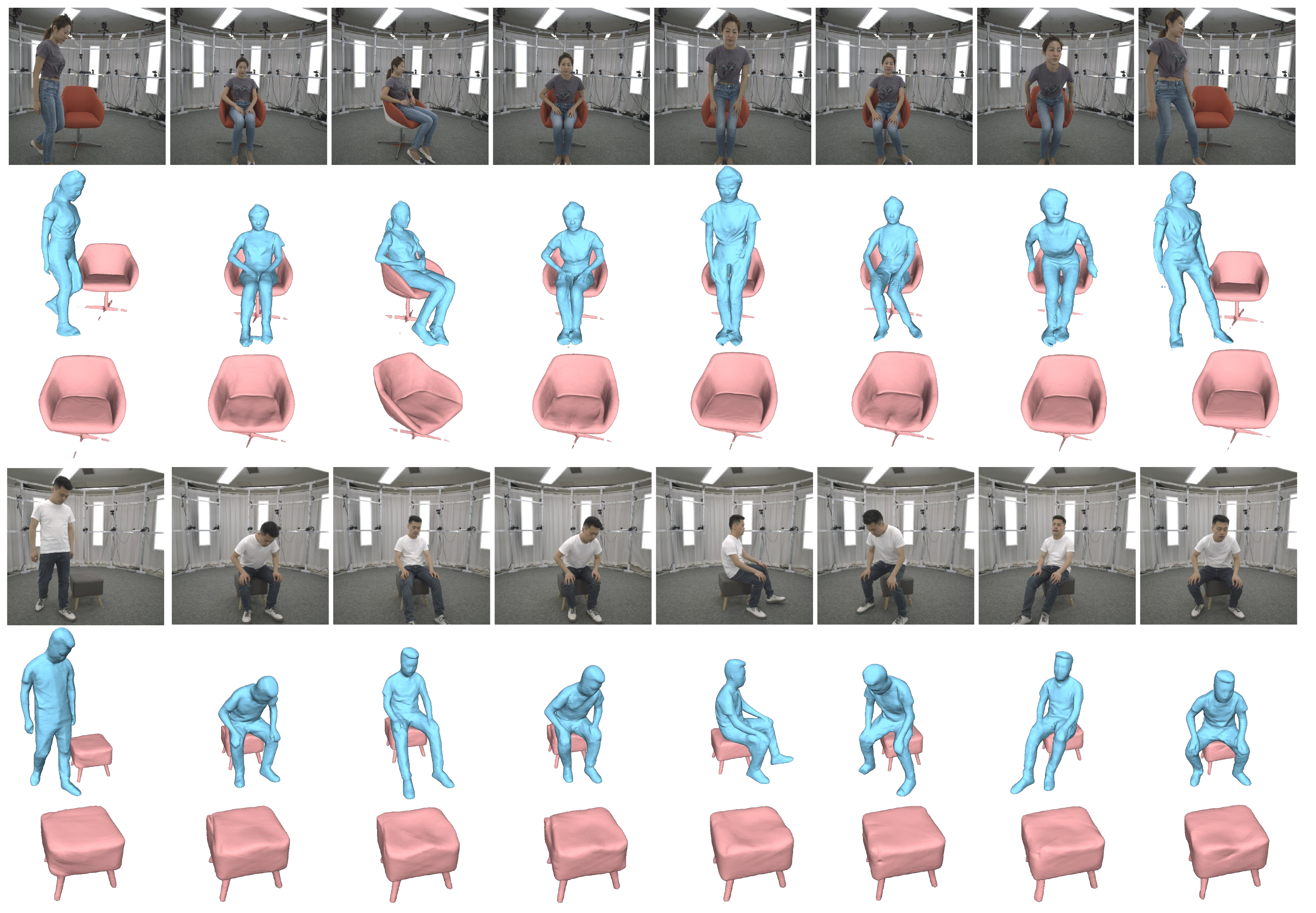}
   \caption{Results of Ins-HOI on real-world data under 6-view input settings. We showcases 1 of 6 inputs, instance-level reconstructions, and the surface deformation of the chairs as the interaction progresses. Note that some cases of fragment reconstruction are due to image segmentation.}
   \label{fig:realworld}
\end{figure*}

Although our model is trained on virtual multi-view rendering images of 3D models, the high fidelity scan in our dataset ensures that there is minimal difference from real-world images. This enable us to further validate our model in real-world scenarios. We use 6 calibrated cameras evenly arranged around a circle to capture human interacting with chair. The collected images are processed through SAM~\cite{kirillov2023segany} to extract human and chair as foreground, serving as inputs of our model. We select key frames from video sequence to present an evaluation of our model's performance in dynamic human-chair interaction scenarios in Fig.~\ref{fig:realworld}. Our results showcase the smooth transition when a person gradually sit down or stand up from a chair. During these movements, our model captures the chair's deformation—from being compressed under weight to returning to its normal state—with remarkable smoothness and realism. This continuous and realistic deformation not only highlights the precision of our method but also underscores its potential for accurately simulating dynamic interactions in real-world scenarios. Besides the plausible results for invisible contact surfaces, our instance-level reconstructions of human and chair are also in high quality and continuity. Due to the inherent limitations of paper format, we can only showcase static results within the paper. For a comprehensive view of continuous and dynamic effects, we kindly refer readers to our project page.

\section{Discussion}\label{sec:conclusion}
\paragraph{Conclusion. }
Modeling human/hand-object interactions accurately and robustly is an exceedingly complex task due to factors such as occlusion and contact. 
In contrast to previous solutions that rely on mocap and template representations, we advocate for capturing the intricate geometry of these interactions through implicit surface representations, using sparse-view inputs.
To this end, we present Ins-HOI, an innovative end-to-end framework for instance-level reconstruction of human/hand-object interactions.
This can not only allow us to generally reconstruct instance-level implicit functions for human/hand and objects, but also learn continuous reconstruction with soft deformations on the invisible contact surfaces. 
To facilitate the research, we have curated two comprehensive 3D scan datasets, capturing the most commonly occurring interactions in real-world: human-chair and hand-object interaction. 
We have conducted extensive experiments and established a robust benchmark that validates the reconstruction results of our Ins-HOI, showcasing the effectiveness and generalizability of our methods. 
From a broader perspective, our method eliminates the need for a heavy capture system, elegantly relying on sparse view RGB inputs. 
By leveraging our pre-trained model, it enables individual reconstruction of interacting instances and can be easily deployed in real-world scenarios. 
We believe that our method paves the way for a new research direction in modeling intricate human-object interactions. 

\paragraph{Limitation and Future Work. }
Although our method is capable of reconstructing invisible contact regions, and we have validated it through both qualitative and quantitative assessments, it is important to acknowledge that our results, while reasonable, are approximations of the ground truth. Capturing the precise deformations of interaction surfaces accurately remains a significant challenge currently. 
Furthermore, our approach requires slight fine-tuning when introduced to a novel object type. This can be addressed by incorporating a wider variety of objects into training dataset or integrating object as priors into the network. Lastly, although our experiments focus on interactions between two instances, the design of our method inherently supports extension to multiple objects or scene. 

In future work, our method and dataset could be integrated with newer 3D representations, such as generalizable NeRF~\cite{mildenhall2020nerf} and Gaussian splatting~\cite{kerbl3Dgaussians}, which aims to reconstruct textures, reduce the number of input viewpoints, and achieve better geometry for more efficient instance-level reconstruction.

\bibliographystyle{IEEEtran}
\bibliography{IEEEabrv,bibtex}

\vfill

\end{document}